\documentclass[11pt]{article}%
\usepackage{amsmath}
\usepackage{amsfonts}
\usepackage{amssymb}
\usepackage{graphicx}%
\providecommand{\U}[1]{\protect \rule{.1in}{.1in}}

\begin{document}

\title{Generating Survival Interpretable Trajectories and Data}
\author{Andrei V. Konstantinov, Stanislav R. Kirpichenko, and Lev V. Utkin\\{\small {Higher School of Artificial Intelligence Technologies}}\\{\small {Peter the Great St.Petersburg Polytechnic University}}\\{\small {St.Petersburg, Russia}}\\{\small {e-mail: andrue.konst@gmail.com, kirpichenko.sr@edu.spbstu.ru,
lev.utkin@gmail.com}}}
\date{}
\maketitle

\begin{abstract}
A new model for generating survival trajectories and data based on applying an
autoencoder of a specific structure is proposed. It solves three tasks. First,
it provides predictions in the form of the expected event time and the
survival function for a new generated feature vector on the basis of the Beran
estimator. Second, the model generates additional data based on a given
training set that would supplement the original dataset. Third, the most
important, it generates a prototype time-dependent trajectory for an object,
which characterizes how features of the object could be changed to achieve a
different time to an event. The trajectory can be viewed as a type of the
counterfactual explanation. The proposed model is robust during training and
inference due to a specific weighting scheme incorporating into the
variational autoencoder. The model also determines the censored indicators of
new generated data by solving a classification task. The paper demonstrates
the efficiency and properties of the proposed model using numerical
experiments on synthetic and real datasets. The code of the algorithm
implementing the proposed model is publicly available.


\end{abstract}

\section{Introduction}

There are many applications, including medicine, reliability, safety, finance,
with problems handling time-to-event data. The related problems are often
solved within the context of survival analysis
\cite{Hosmer-Lemeshow-May-2008,Wang-Li-Reddy-2019} which considers two types
of observations: censored and uncensored. A censored observation takes place
when we do not observe the corresponding event because it occurs after the
observation. When we observe the event, then the corresponding observation is
uncensored. The censored and uncensored observations can be regarded as one of
the main challenges in survival analysis.

Many survival models have been developed to deal with censored and uncensored
data in the context of survival analysis
\cite{Wang-Li-Reddy-2019,EmmertStreib-Dehmer-19,Ranganath-etal-2016,salerno2023high,Wiegrebe-etal-23}%
. The models solve the classification and regression tasks under various
conditions and constraints imposed on the data within a particular
application. However, the most models require a large amount of training data
to provide accurate predictions. One of the ways to overcome this problem is
to generate synthetic data. Due to peculiarities of survival samples, for
example, due to their censoring, there are a few methods for the data
generation. Most methods generate survival times to simulate the Cox
proportional hazards model \cite{Cox-1972}. Bender et al.
\cite{Bender-etal-2005} show how the exponential, the Weibull and the Gompertz
distributions can be applied to generate appropriate survival times for
simulation studies. The authors propose relationships between the survival
time and the hazard function of the Cox models using the above probability
distributions of time-to-event, which links the survival time with a feature
vector characterizing an object of interest. Austin \cite{Austin-12} extends
the approach for generating survival times proposed in \cite{Bender-etal-2005}
to the case of time-varying covariates. Extensions of methods for generating
survival times have been also proposed in
\cite{harden2019simulating,hendry2014data,montez2017guidelines,Ngwa-etal-22}.
Reviews of the algorithms for generating survival data can be found in
\cite{montez2017guidelines,sylvestre2008comparison}. However, the presented
results remain in the framework of Cox models.

A quite different generative model handling survival data and called
SurvivalGAN was proposed by Norcliffe et al. \cite{norcliffe2023survivalgan}.
SurvivalGAN goes beyond the Cox model and generates synthetic survival data
from any probability distribution that the corresponding training set may
have. It efficiently takes into account a structure of the training set that
is the relative location of instances in the dataset. SurvivalGAN is a
powerful and outstanding tool for generating survival data. However, it
requires that a censored indicator be specified in advance to generate the
event time. If the user specifies as a condition that a generated instance is
uncensored, but the instance is located in an area of censored data, then the
model may provide incorrect results.

We propose a new model for generating survival data based on applying a
variational autoencoder (VAE) \cite{Kingma-Welling-2014}. Its main aim is to
generate a time-dependent \emph{trajectory} of an object, which answers the
following question: \emph{What features of the object should be changed and
how so that the corresponding event time would be different, for example,
longer?} The trajectory is a set of feature vectors depending on time. It can
be viewed as a type of the counterfactual explanation
\cite{Guidotti-etal-2019a,Sokol-Flach-2019,Wachter-etal-2017} which describes
the smallest change to the feature values that changes a prediction to a
predefined output \cite{Molnar-2019}. Suppose that we have a dataset of
patients with a certain disease such that feature vectors are various
combinations of drugs given to patients. It is known that a patient from the
dataset is treated with a specific combination of drugs, and the patient's
recovery time is predicted to be one month. By constructing the patient's
trajectory, we can determine how to change the combination of drugs to reduce
the recovery time till three weeks.

An important feature of the proposed model is its robustness both during
training and during generating new data (inference). For each time and for
each feature vector, a set of close embeddings is generated so that their
weighted average determines the generated trajectory. The generated set of
feature vectors can be regarded as the noise incorporating into the training
and inference processes to ensure robustness. In addition to the trajectory
for a new feature vector or a feature vector from the dataset, the model
generates a random event time and an expected event time. It allows us to
predict survival characteristics, including the survival function (SF) like a
conventional machine learning model. Another important feature of the proposed
model is that the censored indicator, which is generated in many models by
using the Bernoulli distribution, is determined by solving a classification
task. For this purpose, a binary classifier is trained on the available
dataset such that each instance for the classifier consists of the
concatenated original feature vectors and the corresponding the event times,
but the target value is nothing else but the censored indicator.

A scheme of the proposed autoencoder architecture is depicted in Fig.
\ref{f:survival vae}, and it is trained in the end-to-end manner.

In sum, the contribution of the paper can be formulated as follows:

\begin{enumerate}
\item A new model for generating survival data based on applying the VAE is
proposed. It generates the \emph{prototype time-dependent trajectory} which
characterizes how features of an object could be changed by different times to
event of interest. For each feature vector $\mathbf{x}$, the trajectory
traverses the point $(\mathbf{x},\mathbb{E}[t|\mathbf{x}])$ in the scenario or
at least be close to it.

\item The proposed model solves the survival task, i.e., for a new feature
vector, the model provides predictions in the form of the expected time to
event and the SF.

\item The model generates additional data based on a given training set that
would supplement the original dataset. We consider the \emph{conditional
generation} which means that, given some input vector $\mathbf{x}$, the model
generates the output points close to $\mathbf{x}$.
\end{enumerate}

Several numerical experiments with the proposed model on synthetic and real
datasets demonstrate its efficiency and properties. The code of the algorithm
implementing the model can be found at https://github.com/NTAILab/SurvTraj.

The paper is organized as follows. Concepts of survival analysis, including
SFs, C-index, the Cox model and the Beran estimator are introduced in Section
2. A detailed description of the proposed model is provided in Section 3.
Numerical experiments with synthetic data and real data are given in Section
4. Concluding remarks can be found in Section 5.

\section{Concepts of survival analysis}

An instance (object) in survival analysis is usually represented by a triplet
$(\mathbf{x}_{i},\delta_{i},T_{i})$, where $\mathbf{x}_{i}^{\mathrm{T}%
}=(x_{i1},...,x_{id})$ is the vector of the instance features; $T_{i}$ is time
to event of interest for the $i$-th instance. If the event of interest is
observed, then $T_{i}$ is the time between a baseline time and the time of
event happening. In this case, an uncensored observation takes place and
$\delta_{i}=1$. Another case is when the event of interest is not observed.
Then $T_{i}$ is the time between the baseline time and the end of the
observation. In this case, a censored observation takes place and $\delta
_{i}=0$. There are different types of censored observations. We will consider
only right-censoring, where the observed survival time is less than or equal
to the true survival time \cite{Hosmer-Lemeshow-May-2008}. Given a training
set $\mathcal{A}$ consisting of $n$ triplets $(\mathbf{x}_{i},\delta_{i}%
,T_{i})$, $i=1,...,n$, the goal of survival analysis is to estimate the time
to the event of interest $T$ for a new instance $\mathbf{x}$ by using
$\mathcal{A}$.

Key concepts in survival analysis are SFs $S(t\mid \mathbf{x})$ and hazard
functions $h(t\mid \mathbf{x})$, which describe probability distributions of
the event times. The SF is the probability of surviving up to time $t$, that
is $S(t\mid \mathbf{x})=\Pr \{T>t|\mathbf{x}\}$. The hazard function
$h(t\mid \mathbf{x})$ is the rate of the event at time $t$ given that no event
occurred before time $t$. The hazard function can be expressed through the SF
as follows \cite{Hosmer-Lemeshow-May-2008}:
\begin{equation}
h(t\mid \mathbf{x})=-\frac{\mathrm{d}}{\mathrm{d}t}\ln S(t\mid \mathbf{x}).
\end{equation}

One of the measures to compare survival models is the C-index proposed by
Harrell et al. \cite{Harrell-etal-1982}. It estimates the probability that the
event times of a pair of instances are correctly ranking. Different forms of
the C-index can be found in literature. We use one of the forms proposed in
\cite{Uno-etal-11}:
\begin{equation}
C=\frac{\sum \nolimits_{i,j}\mathbb{I}[T_{i}<T_{j}]\cdot \mathbb{I}[\widehat
{T}_{i}<\widehat{T}_{j}]\cdot \delta_{i}}{\sum \nolimits_{i,j}\mathbb{I}%
[T_{i}<T_{j}]\cdot \delta_{i}},
\end{equation}
where $\widehat{T}_{i}$ and $\widehat{T}_{j}$ are the predicted survival
durations; $\mathbb{I}[\cdot]$ is the indicator function.

The next concept of survival analysis is the Cox proportional hazards model.
According to the model, the hazard function at time $t$ given vector
$\mathbf{x}$ is defined as \cite{Cox-1972,Hosmer-Lemeshow-May-2008}:
\begin{equation}
h(t\mid \mathbf{x},\mathbf{b})=h_{0}(t)\exp \left(  \mathbf{b}^{\mathrm{T}%
}\mathbf{x}\right)  . \label{SurvLIME1_10}%
\end{equation}

Here $h_{0}(t)$ is a baseline hazard function which does not depend on the
vector $\mathbf{x}$ and the vector $\mathbf{b}$; $\mathbf{b}^{\mathrm{T}%
}=(b_{1},...,b_{m})$ is a vector of the unknown regression coefficients or the
model parameters. The baseline hazard function represents the hazard when all
of the covariates are equal to zero.

The SF in the framework of the Cox model is
\begin{equation}
S(t\mid \mathbf{x},\mathbf{b})=\left(  S_{0}(t)\right)  ^{\exp \left(
\mathbf{b}^{\mathrm{T}}\mathbf{x}\right)  }, \label{Cox_SF}%
\end{equation}
where $S_{0}(t)$ is the baseline SF.

Another important model is the Beran estimator. Given the dataset
$\mathcal{A}$, the SF can be estimated by using the Beran estimator
\cite{Beran-81} as follows:
\begin{equation}
S(t\mid \mathbf{x},\mathcal{A})=\prod_{T_{i}\leq t}\left \{  1-\frac
{W(\mathbf{x},\mathbf{x}_{i})}{1-\sum_{j=1}^{i-1}W(\mathbf{x},\mathbf{x}_{j}%
)}\right \}  ^{\delta_{i}}, \label{Beran_est}%
\end{equation}
where time moments are ordered; the weight $W(\mathbf{x},\mathbf{x}_{i})$
conforms with relevance of the $i$-th instance $\mathbf{x}_{i}$ to the vector
$\mathbf{x}$ and can be defined through kernels as%
\begin{equation}
W(\mathbf{x},\mathbf{x}_{i})=\frac{K(\mathbf{x},\mathbf{x}_{i})}{\sum
_{j=1}^{n}K(\mathbf{x},\mathbf{x}_{j})}.
\end{equation}

If we use the Gaussian kernel, then the weights $W(\mathbf{x},\mathbf{x}_{i})$
are of the form:
\begin{equation}
W(\mathbf{x},\mathbf{x}_{i})=\text{\textrm{softmax}}\left(  -\frac{\left \Vert
\mathbf{x}-\mathbf{x}_{i}\right \Vert ^{2}}{\tau}\right)  , \label{softmax}%
\end{equation}
where $\tau$ is a temperature parameter.

The Beran estimator is trained on the dataset $\mathcal{A}$ and is used for
new objects $\mathbf{x}$. It can be regarded as a generalization of the
Kaplan-Meier estimator \cite{Wang-Li-Reddy-2019} because it is reduced to the
Kaplan-Meier estimator if the weights $W(\mathbf{x},\mathbf{x}_{i})$ take
values $W(\mathbf{x},\mathbf{x}_{i})=1/n$ for all $i=1,...,n$.

\section{Generating trajectories and data}

An idea for constructing the time trajectory for an object $\mathbf{x}$ is to
apply a VAE which

\begin{itemize}
\item is trained on subsets $\mathcal{A}_{r}$ of $r$ training instances
$\widetilde{\mathbf{x}}_{1},...\widetilde{\mathbf{x}}_{r}$ by computing the
corresponding random embeddings $\mu(\widetilde{\mathbf{x}}_{1}),...\mu
(\widetilde{\mathbf{x}}_{r})$, which are used to learn the survival model (the
Beran estimator);

\item generates a set of $m$ embeddings $\mathbf{z}_{1},...,\mathbf{z}_{m}$
for each feature vector $\mathbf{x}$ by means of the encoder;

\item learns the Beran estimator for computing SFs for embeddings, for
computing the expected event time $\widehat{T}$, for generating a new time to
event $T_{gen}$, and for computing loss functions to learn the whole model;

\item computes a \emph{prototype time trajectory} $\xi_{\mathbf{z}}(t)$ at
time moments $t_{1},...,t_{v}$ by using the generated embeddings
$\mathbf{z}_{j}$;

\item then uses the decoder to obtain the reconstructed trajectory
$\xi_{\mathbf{x}}(t)$ for $\mathbf{x}$.
\end{itemize}

The detailed architecture and peculiarities of the VAE will be considered
later. We apply the Wasserstein autoencoder \cite{tolstikhin2017wasserstein}
as a basis for constructing the model generating the trajectory and
implementing the generation procedures. The Wasserstein autoencoder aims to
generate latent representations that are close to a standard normal
distribution, which can help to improve the performance of tasks. It learns
from a loss function that includes the maximum mean discrepancy regularization.

\begin{figure}
[ptb]
\begin{center}
\includegraphics[
height=2.5346in,
width=6.3805in
]%
{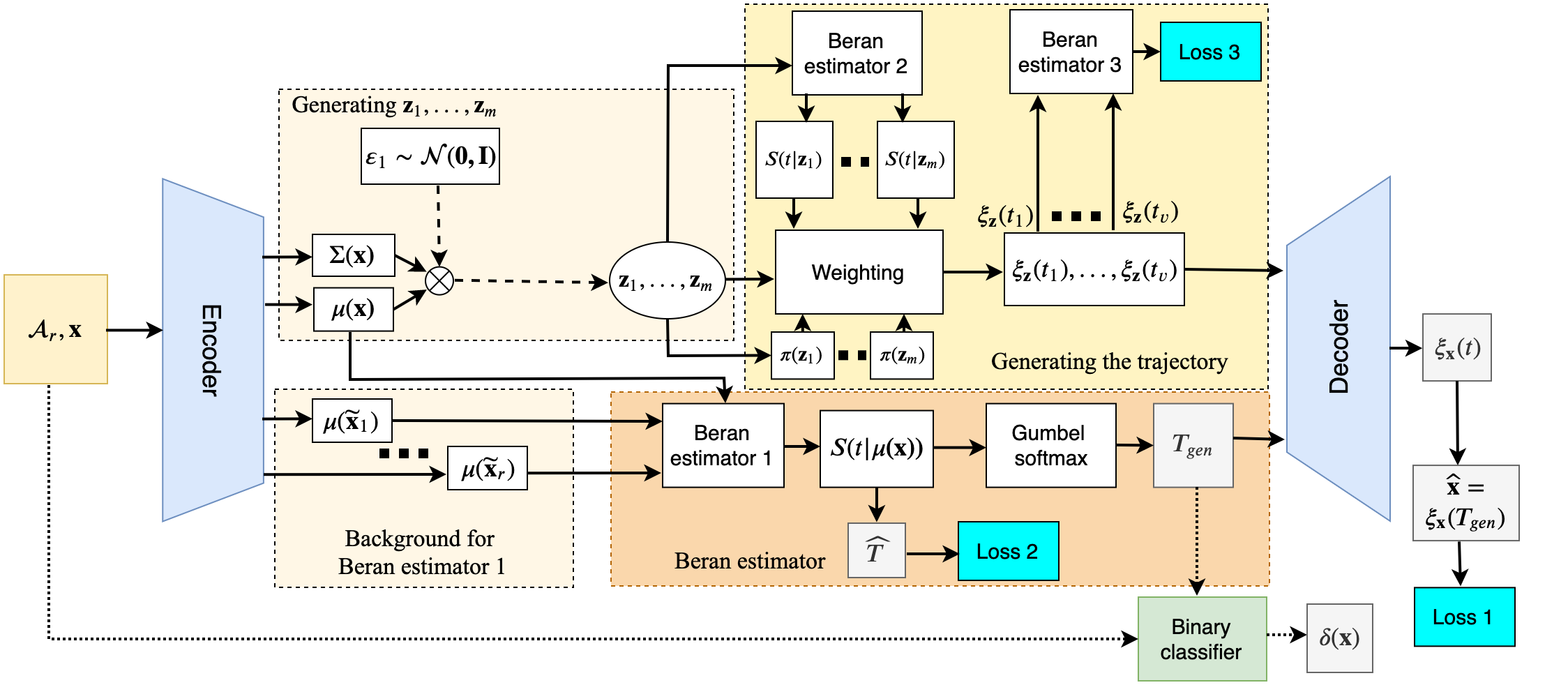}%
\caption{A scheme of the proposed model}%
\label{f:survival vae}%
\end{center}
\end{figure}

A general scheme of the proposed model based on applying the VAE is depicted
in Fig. \ref{f:survival vae}. It serves as a kind of a \textquotedblleft
container\textquotedblright \ that holds all end-to-end trainable parts of the
considered model.%

\subsection{Encoder part and training epochs}

The first part of the VAE is the encoder which provides parameters
$\mu(\mathbf{x})$ and $\Sigma(\mathbf{x})$ for generating embeddings in the
hidden space producing the time trajectory and parameters $\mu(\widetilde
{\mathbf{x}})$ and $\Sigma(\widetilde{\mathbf{x}})$ for generating embeddings
to \textquotedblleft learning\textquotedblright \ the Beran estimator. The
encoder converts input feature vectors $\mathbf{x}$ into the hidden space $Z$.
According to the standard VAE, the mapping is performed using the
\textquotedblleft reparametrization trick\textquotedblright. Each training
epoch includes solving $M$ tasks such that each task consists of the following
set:%
\begin{equation}
\{ \underbrace{(\widetilde{\mathbf{x}}_{i},T_{i},\delta_{i}),i=1,...,r}%
_{\text{background in the Beran estimator}},\underbrace{\mathbf{x,}T,\delta
}_{\text{input instance}}\}. \label{train_epoch}%
\end{equation}

The training dataset in this case contains the following triplets:
$(\widetilde{\mathbf{x}}_{i},T_{i},\delta_{i})$, $i=1,...,r$, which form the
datasubset $\mathcal{A}_{r}\subset \mathcal{A}$, and the triplet $(\mathbf{x}%
,T,\delta)$ which is taken from the dataset $\mathcal{A}\backslash
\mathcal{A}_{r}$ during training and is a new instance during inference. Here
$r$ is a hyperparameter. In order to differ the points from $\mathcal{A}_{r}$
and $\mathcal{A}\backslash \mathcal{A}_{r}$, we denote the selected feature
vectors in $\mathcal{A}_{r}$ as $\widetilde{\mathbf{x}}$. Thus, $M$ sets
$\mathcal{A}_{r}$ of $r$ points are selected on each epoch, which are used as
the training set, and the remaining $n-r$ points are processed through the
model directly and are passed to the loss function, after which the
optimization is performed by the error backpropagation. After training the
model on several epochs, the background for the Beran estimator is set to the
entire training set.

\begin{figure}
[ptb]
\begin{center}
\includegraphics[
height=1.3422in,
width=2.9698in
]%
{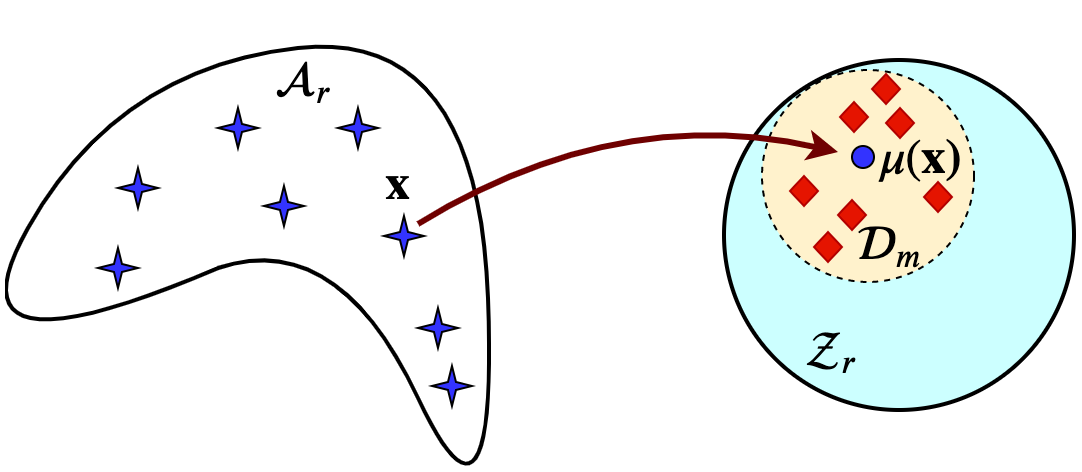}%
\caption{The original set $\mathcal{A}_{r}$ of vectors $\mathbf{x}_{i}$ and
the set $\mathcal{D}_{m}$ of vectors $\widetilde{\mathbf{z}}_{1}%
,...,\widetilde{\mathbf{z}}_{m}$ normally distributed around vector
$\mathbf{z}$}%
\label{f:transform_2}%
\end{center}
\end{figure}

In order to describe the whole scheme of training and using the VAE, we
consider two subsets of vectors generated by the encoder. The first subset
corresponding to the upper path in the scheme in Fig. \ref{f:survival vae}
(Generating $\mathbf{z}_{1},,,.,\mathbf{z}_{m}$) consists of two vectors:
$\mu(\mathbf{x})$ (mean values) and $\Sigma(\mathbf{x})$ (standard
deviations). It should be noted that we consider $\Sigma$ as a vector, but not
as a covariance matrix, because we aim to get uncorrelated features in the
embedding space $\mathcal{Z}$ of the VAE. These parameter vectors are used to
generate random vectors $\mathbf{z}_{1},,,.,\mathbf{z}_{m}$ calculated as
$\mathbf{z}_{i}=\mu(\mathbf{x})+\varepsilon_{1}\cdot \Sigma(\mathbf{x})$, where
$\varepsilon_{1}$ is the normally generated vector of noise $\varepsilon
_{1}\sim \mathcal{N}(\mathbf{0},\mathbf{I})$, $\mathbf{0}=(0,...0)$,
$\mathbf{I}=(1,...1)$. Vectors $\mathbf{z}_{1},,,.,\mathbf{z}_{m}$ are used
for training as well as for inference. They are located around $\mu
(\mathbf{x})$ and form a set $\mathcal{D}_{m}$ of normally distributed points,
which is schematically shown in Fig. \ref{f:transform_2}. The set
$\mathcal{D}_{m}$ is used to compute the robust trajectory $\xi_{\mathbf{z}%
}(t)$.%

The second subset of vectors generated by the encoder consists of the
functions $\mu(\widetilde{\mathbf{x}}_{1}),...\mu(\widetilde{\mathbf{x}}_{r})$
of the vectors $\widetilde{\mathbf{x}}_{1},...,\widetilde{\mathbf{x}}_{r}$
from $\mathcal{A}_{r}$. In this case, the set $\mathcal{A}_{r}$ is selected
from the entire dataset to learn the Beran estimator which is used to predict
the SF and the expected event time for the vector $\mu(\mathbf{x})$.
Therefore, vectors $\mu(\widetilde{\mathbf{x}}_{1}),...\mu(\widetilde
{\mathbf{x}}_{r})$ can be regarded as the background for the Beran estimator.
In contrast to vectors $\mathbf{z}_{1},,,.,\mathbf{z}_{m}$ which are generated
for the feature vector $\mathbf{x}$, each vector $\mu(\widetilde{\mathbf{x}%
}_{i})$ is generated for the vector $\widetilde{\mathbf{x}}_{i}$ from
$\mathcal{A}_{r}$. The second pair corresponds to the bottom path in the
scheme in Fig. \ref{f:survival vae} (Background for Beran estimator 1).

\subsection{The prototype embedding trajectory}

Let us consider how to use the trained survival model (the Beran estimator) to
compute the SF $S(t\mid \mathbf{z})$ and the trajectory $\xi_{\mathbf{z}}(t)$
(see Generating trajectory in Fig. \ref{f:survival vae}).

Let $0<t_{1}<...<t_{n}$ be the distinct times to event of interest from the
set $\{T_{1},...,T_{n}\}$, where $t_{1}=\min_{k=1,...,n}T_{k}$ and $t_{n}%
=\max_{k=1,...,n}T_{k}$. Suppose that a new vector $\mathbf{x}$ is fed to the
encoder of the VAE. The encoder produces vectors $\mu(\mathbf{x})$ and
$\Sigma(\mathbf{x})$. In accordance with these parameters and the random noise
$\varepsilon_{1}$, the random embeddings $\mathbf{z}_{1},...,\mathbf{z}_{m}$
are generated from the normal distribution $\mathcal{N}(\mu(\mathbf{x}%
),\Sigma(\mathbf{x}))$. For every $\mathbf{z}_{i}$ from $\mathcal{D}_{m}$, we
can find the density function $\pi(t\mid \mathbf{z}_{i})$ by using the trained
survival model predicting the SF $S(t\mid \mathbf{z}_{i})$. The density
function can be expressed through the SF $S(t\mid \mathbf{z}_{i})$ as:
\begin{equation}
\pi(t\mid \mathbf{z}_{i})=-\frac{\mathrm{d}S(t\mid \mathbf{z}_{i})}{\mathrm{d}%
t}.
\end{equation}

However, our goal is to find another density function $\pi(\mathbf{z}_{i}\mid
t)$ which allows us to generate the vectors $\mathbf{z}_{i}$ at different time
moments. The density $\pi(\mathbf{z}_{i}\mid t)$ can be computed by using the
Bayes rule:%

\begin{equation}
\pi(\mathbf{z}_{i}\mid t)=\dfrac{\pi(t\mid \mathbf{z}_{i})\cdot \pi
(\mathbf{z}_{i})}{\pi(t)}. \label{bayes_rule}%
\end{equation}

Here $\pi(\mathbf{z}_{i})$ is a priori density which can be estimated by
applying several ways, for example, by means of the kernel density estimator.
The density $\pi(t)$ can be estimated by using the Kaplan-Meier estimator.
However, we do not need to estimate it because it can be regarded as a
normalizing coefficient.

Now we have everything to compute $\pi(\mathbf{z}_{i}\mid t)$ and can consider
how to use it to generate new points in accordance with this density.

Let us introduce a \emph{prototype embedding trajectory} $\xi_{\mathbf{z}}(t)$
taking a value at each time $t$ as a mean value of vectors $\mathbf{z}_{i}$,
$i=1,...,m$, with respect to densities $\pi(\mathbf{z}_{i}\mid t)$,
$i=1,...,m$, as follows:%
\begin{equation}
\xi_{\mathbf{z}}(t)=\sum_{i=1}^{m}\frac{\pi(\mathbf{z}_{i}\mid t)\cdot
\mathbf{z}_{i}}{\sum_{j=1}^{m}\pi(\mathbf{z}_{j}\mid t)}. \label{trajectory_1}%
\end{equation}

After substituting the Bayes rule (\ref{bayes_rule}) into the expression for
the trajectory (\ref{trajectory_1}), we obtain
\begin{equation}
\xi_{\mathbf{z}}(t)=\sum_{i=1}^{m}\frac{\pi(t\mid \mathbf{z}_{i})\cdot
\pi(\mathbf{z}_{i})\cdot \mathbf{z}_{i}}{\sum_{j=1}^{m}\pi(t\mid \mathbf{z}%
_{j})\cdot \pi(\mathbf{z}_{j})}=\sum_{i=1}^{m}\alpha_{i}(t)\cdot \mathbf{z}_{i},
\label{main_express}%
\end{equation}
where $\alpha_{i}(t)$ is a normalized weight of each $\mathbf{z}_{i}$ in the
trajectory at time $t$, which defined as
\begin{equation}
\alpha_{i}(t)=\frac{\pi(t\mid \mathbf{z}_{i})\cdot \pi(\mathbf{z}_{i})}%
{\sum_{j=1}^{m}\pi(t\mid \mathbf{z}_{j})\cdot \pi(\mathbf{z}_{j})}.
\label{alfa1}%
\end{equation}

It can be seen from (\ref{main_express}) that the trajectory $\xi_{\mathbf{z}%
}(t)$ is the weighted sum of generated vectors $\mathbf{z}_{i}$, $i=1,...,m$,
depicted in Fig. \ref{f:survival vae} as the block \textquotedblleft
Weighting\textquotedblright. As a result, we obtain the robust trajectory for
the latent representation $\mathbf{z}$ or $\mu(\mathbf{x})$.

Let us consider how to compute the density $\pi(t\mid \mathbf{z}_{i})$ in
accordance with the Beran estimator (Beran estimator 2 in Fig.
\ref{f:survival vae}). First, the SF $S(t\mid \mathbf{z}_{i})$ is determined by
using (\ref{Beran_est}) as:
\begin{equation}
S(t\mid \mathbf{z}_{i})=\prod_{\widetilde{T}_{i}\leq t}\left \{  1-\frac
{W(\mathbf{z}_{i},\mu(\widetilde{\mathbf{x}}_{i}))}{1-\sum_{j=1}%
^{i-1}W(\mathbf{z}_{j},\mu(\widetilde{\mathbf{x}}_{j}))}\right \}  ^{\delta
_{i}},
\end{equation}
where $\widetilde{T}_{i}$ is the event time corresponding to the vector
$\widetilde{\mathbf{x}}_{i}$ from $\mathcal{A}_{r}$.

Second, due to the final number of training instances, the Beran estimator
provides a step-wise SF represented as follows:
\begin{equation}
S(t\mid \mathbf{z}_{i})=\sum \limits_{j=0}^{n-1}S_{j}\cdot \mathbb{I}%
\{t\in \lbrack t_{j},t_{j+1})\}, \label{Beran_SF}%
\end{equation}
where $S_{j}=S(t_{j}\mid \mathbf{z}_{i})$ is the SF in the time interval
$[t_{j},t_{j+1}]$ obtained from (\ref{Beran_est}); $S_{0}=1$ by\ $t_{0}=0$;
$\mathbb{I}\{t\in \lbrack t_{j},t_{j+1})\}$ is the indicator function taking
the value $1$ if $t\in \lbrack t_{j},t_{j+1})$, and $0$, otherwise.

The probability density function $\pi(t\mid \mathbf{z}_{i})$ can be calculated
as:
\begin{equation}
\pi(t|\mathbf{z}_{i})=\sum_{j=0}^{n-1}\left(  S_{j}-S_{j+1}\right)
\cdot \mathbb{\delta}\{t=t_{j}\}, \label{Beran_density}%
\end{equation}
where $\mathbb{\delta}\{t=t_{j}\}$ is the Dirac delta function.

Let us replace the density $\pi(t\mid \mathbf{z}_{i})$ with the discrete
probability distribution $\left(  p(t_{1}\mid \mathbf{z}_{i}),...,p(t_{n}%
\mid \mathbf{z}_{i})\right)  $ such that $p(t_{j}\mid \mathbf{z}_{i}%
)=S_{j-1}-S_{j}$. Then (\ref{alfa1}) can be represented in another form:
\begin{equation}
\alpha_{i}(t_{j})=\frac{p(t_{j}\mid \mathbf{z}_{i})\pi(\mathbf{z}_{i})}%
{\sum_{l=1}^{m}p(t_{j}\mid \mathbf{z}_{l})\cdot \pi(\mathbf{z}_{l}%
)},\ j=1,...,n.
\end{equation}

Since coefficients $\alpha_{1},...,\alpha_{m}$ are normalized, then they form
the convex combination of $\mathbf{z}_{1},...,\mathbf{z}_{m}$.

Vectors $\mathbf{z}_{i}$ are governed by the normal distribution
$\mathcal{N}(\mu(\mathbf{x}),\Sigma(\mathbf{x}))$, therefore, the density
$\pi(\mathbf{z}_{i})$ is determined as
\begin{equation}
\pi(\mathbf{z}_{i})\varpropto \exp \left(  -\frac{1}{2}(\mathbf{z}_{i}%
-\mu(\mathbf{x}))^{\top}\left(  \Sigma(\mathbf{x})\right)  ^{-1}%
(\mathbf{z}_{i}-\mu(\mathbf{x}))\right)  . \label{norm_density}%
\end{equation}

It is important to point out that $p(t\mid \mathbf{z}_{i})$ as well as
$\pi(t\mid \mathbf{z}_{i})$ are defined at time points $t_{1},...,t_{n}$.
However, when the trajectory is constructed, it is necessary to ensure that
the model takes into account the entire context. To cope with this difficulty,
we propose to smooth the density function $\pi(t\mid \mathbf{z}_{i})$ to obtain
a smooth trajectory. The smoothing is carried out using a convex combination
with coefficients $\{ \beta_{1}(t),...,\beta_{n}(t)\}$ determined by means of
the softmin operation with respect to the distance from $t$ to $\{t_{1}%
,...,t_{n}\}$, respectively. Then we can write for the smooth version of
$\pi(t\mid \mathbf{z}_{i})$ denoted as $\widetilde{\pi}(t\mid \mathbf{z}_{i})$:
\begin{equation}
\widetilde{\pi}(t\mid \mathbf{z}_{i})=\sum \limits_{i=1}^{n}\beta_{i}(t)\cdot
\pi(t_{i}\mid \mathbf{z}_{i}),
\end{equation}
where
\begin{equation}
\beta_{i}(t)=\mathrm{softmin}(\eta \cdot|t-t_{i}|),\ i=1,...,n,
\end{equation}
$\eta$ is a training parameter; $\mathrm{softmin}(x)=\mathrm{softmax}(-x)$.

The trajectory $\xi_{\mathbf{z}}(t)$ is determined for the finite set of time
moments $t_{1},...,t_{v}$ which are selected as follows: $t_{k}=t_{k-1}%
+(t_{\max}-t_{\min})/v$, where $t_{\min}$, $t_{\max}$ are the smallest and the
largest times to event from the training set, $t_{0}=t_{\min}$, $k=1,...,v$.

In order to compute the corresponding \emph{prototype trajectory}
$\xi_{\mathbf{x}}(t)$ for the vector $\mathbf{x}$, we use the decoder of the
VAE. The prototype trajectory $\xi_{\mathbf{x}}(t)$ at each time moment can be
viewed as some points in the dataset domain, i.e., for each time $t_{j}$, we
can construct a point (vector) $\xi_{\mathbf{x}}(t_{j})\in \mathbb{R}^{d}$ at
the trajectory $\xi_{\mathbf{x}}(t)$ . The trajectory means which features
should be changed in $\mathbf{x}$ to achieve a certain time $t$ to event.

\subsection{New data generation and the censored indicator}

Another task which can be solved in the framework of the proposed model is to
generate a new survival instance in accordance with the available dataset
$\mathcal{A}$. First, we train the Beran estimator (Beran estimator 1 in Fig.
\ref{f:survival vae}) on the set of vectors $\mu(\widetilde{\mathbf{x}}%
_{1}),...\mu(\widetilde{\mathbf{x}}_{r})$. For the vector $\mu(\mathbf{x})$
corresponding to the input vector $\mathbf{x}$, the SF $S(t\mid \mu
(\mathbf{x}))$ can be estimated by using Beran estimator 1 as follows:%
\begin{equation}
S(t\mid \mu(\mathbf{x}))=\prod_{\widetilde{T}_{i}\leq t}\left \{  1-\frac
{W(\mu(\mathbf{x}),\mu(\widetilde{\mathbf{x}}_{i}))}{1-\sum_{j=1}^{i-1}%
W(\mu(\mathbf{x}),\mu(\widetilde{\mathbf{x}}_{j}))}\right \}  ^{\delta_{i}}.
\end{equation}

Here $\widetilde{T}_{i}$ is the event time corresponding to the vector
$\widetilde{\mathbf{x}}_{i}$ from $\mathcal{A}_{r}$. Hence, a new time
$T_{gen}$ is generated in accordance with $S(t\mid \mu(\mathbf{x}))$ by
applying the Gumbel sampling which has already been used in autoencoders
\cite{jang2016categorical}.

By having the reconstructed trajectory $\xi_{\mathbf{x}}(t)$ and the time
$T_{gen}$, we generate a feature vector $\widehat{\mathbf{x}}=\xi_{\mathbf{x}%
}(T_{gen})$ and write a new instance $(\widehat{\mathbf{x}},T_{gen})$.
However, a complete description of the instance requires to determine the
censored indicator $\delta_{gen}$. In order to find $\delta_{gen}$ for
$(\widehat{\mathbf{x}},T_{gen})$, we introduce a binary classifier which
considers each pair $(\mathbf{x}_{i},T_{i})$ from the training set as a single
feature vector, but $\delta_{i}$ from the training set as a class label taking
values $0$ (a censored event) and $1$ (an uncensored event). If the binary
classifier is trained on the training set $((\mathbf{x}_{i},T_{i}),\delta
_{i})$, $i=1,...,n$, then $\delta_{gen}$ can be predicted on the basis of the
feature vector $(\widehat{\mathbf{x}},T_{gen})$. Finally, we obtain the
triplet $(\widehat{\mathbf{x}},T_{gen},\delta_{gen})$. If the classifier
predicts probabilities of two classes, then the Bernoulli distribution is
applied to generate $\delta_{gen}$. It is important to note that the binary
classifier is trained separately from the VAE training.

It should be noted that the SF $S(t\mid \mu(\mathbf{x}))$ in (\ref{Beran_SF})
is also used for computing the expected time $\widehat{T}$ to event which is
of the form:%
\begin{equation}
\widehat{T}=\sum \limits_{i=0}^{n-1}S_{i}\cdot(t_{i+1}-t_{i}).
\label{Mean_time}%
\end{equation}

The expected time is required for its use in the loss function $\mathcal{L}%
_{\text{Beran}}$ (\textquotedblleft Loss 2\textquotedblright \ in Fig.
\ref{f:survival vae}), which is considered below.

\subsection{Decoder part}

The decoder converts the trajectory $\xi_{\mathbf{z}}(t)$ into the trajectory
$\xi_{\mathbf{x}}(t)$. It also produces the vector $\widehat{\mathbf{x}}$
which is used in the loss function $\mathcal{L}_{\text{WAE}}$. The loss
function is schematically depicted in Fig. \ref{f:survival vae} as
\textquotedblleft Loss 1\textquotedblright.

\subsection{Training the VAE}

The entire model is trained in the end-to-end manner, excluding the part that
generates the censored indicator $\delta_{gen}$. The loss function
$\mathcal{L}$ consists of three parts: the first is responsible for the
accurate estimation of the event times and denoted as $\mathcal{L}%
_{\text{Beran}}$, the second is responsible for the accurate reconstruction
and denoted as $\mathcal{L}_{\text{WAE}}$, the third is for accurate
estimation of the trajectory $\xi_{\mathbf{z}}(t)$ at time moments
$t_{1},...,t_{v}$ denoted as $\mathcal{L}_{\text{Tr}}$. Hence, there holds
\begin{equation}
\mathcal{L}=-\mathcal{L}_{\text{Beran}}+\mathcal{L}_{\text{WAE}}%
-\mathcal{L}_{\text{Tr}}. \label{main_loss}%
\end{equation}

Below $\gamma_{1}$, $\gamma_{2}$, $\gamma_{3}$, and $\gamma_{4}$ are
hyperparameters controlling contributions of the corresponding parts of the
loss function.

The loss function $\mathcal{L}_{\text{Beran}}$ (depicted as \textquotedblleft
Loss 2\textquotedblright \ in Fig. \ref{f:survival vae}) is based on the use of
the C-index softened with a sigmoid function $\sigma$:
\begin{equation}
\mathcal{L}_{\text{Beran}}=\gamma_{1}\frac{\sum_{i,j}\mathbb{I}\{t_{j}<t_{i}\}
\cdot \sigma(\hat{T}_{i}-\hat{T}_{j})\cdot \delta_{j}}{\sum_{i,j}\mathbb{I}%
\{t_{j}<t_{i}\} \cdot \delta_{j}}.
\end{equation}

It is included in $\mathcal{L}$ with minus because $\mathcal{L}_{\text{Beran}%
}$ should be maximized. The temperature parameter $\tau$ of the kernel
(\ref{softmax}) in the Beran estimator is also trained due to $\mathcal{L}%
_{\text{Beran}}$.

The loss function $\mathcal{L}_{\text{WAE}}$ (\textquotedblleft Loss
1\textquotedblright \ in Fig. \ref{f:survival vae}) consists of the mean
squared error on reconstructions and of the regularization $\mathcal{L}%
_{\text{MMD}}$ in the form of the maximum mean discrepancy
\cite{tolstikhin2017wasserstein}:
\begin{equation}
\mathcal{L}_{\text{WAE}}=\frac{\gamma_{2}}{n}\sum \limits_{i=1}^{n}\left \Vert
\mathbf{x}_{i}-\widehat{\mathbf{x}}_{i}\right \Vert ^{2}+\mathcal{L}%
_{\text{MMD}}, \label{loss}%
\end{equation}
where $\widehat{\mathbf{x}}_{1},...,\widehat{\mathbf{x}}_{n}$ are
conditionally generated according to the formula $\widehat{\mathbf{x}}%
=\xi_{\mathbf{x}}(T_{gen})$.

Note that the sets of embeddings $\{ \mathbf{z}_{1}^{(i)},,,.,\mathbf{z}%
_{m}^{(i)}\}$ governed by the the normal distribution $\mathcal{N}%
(\mu(\mathbf{x}_{i}),\Sigma(\mathbf{x}_{i}))$ are generated $n$ times for
every $\mathbf{x}_{i}$, $i=1,...,n$. During training, we take the first
embeddings $\mathbf{z}_{1}^{(i)}$ for all $i=1,...,n$, and compare them with
the embeddings $\widehat{\mathbf{z}}_{i}$ sampled from the normal distribution
$\mathcal{N}(\mathbf{0},\mathbf{1})$. To ensure that all embeddings, including
$\mu(\mathbf{x}_{i})$, are normally distributed, the following regularization
is used:%
\begin{align}
\mathcal{L}_{\text{MMD}}  &  =\frac{\lambda}{n(n-1)}\sum \limits_{l,j=1,l\neq
j}^{n}K(\mathbf{z}_{l},\mathbf{z}_{j})+\frac{\lambda}{n(n-1)}\sum
\limits_{l,j=1,l\neq j}^{n}K(\widehat{\mathbf{z}}_{i},\widehat{\mathbf{z}}%
_{j})\nonumber \\
&  -\frac{2\lambda}{n^{2}}\sum \limits_{l=1}^{n}\sum \limits_{j=1}%
^{n}K(\mathbf{z}_{l},\widehat{\mathbf{z}}_{j}), \label{l_wae}%
\end{align}
where $K(\mathbf{x},\mathbf{y})=C/(C+||\mathbf{x}-\mathbf{y}||_{2}^{2})$ is a
positive-define kernel with the parameter $C=2\cdot \dim(\mathbf{z}{)}$;
$\lambda \geq0$ is a hyperparameter.

The loss function $\mathcal{L}_{\text{Tr}}$ (\textquotedblleft Loss
3\textquotedblright \ in Fig. \ref{f:survival vae}) consists of two parts
$\mathcal{L}_{\text{Tr1}}$ and $\mathcal{L}_{\text{Tr2}}$. The loss function
$\mathcal{L}_{\text{Tr1}}$ is similar to $\mathcal{L}_{\text{Beran}}$, but it
controls how the expected event times $\hat{T}_{1},...,\hat{T}_{v}$ obtained
for elements of the trajectory $\xi_{\mathbf{z}}(t_{1}),...,\xi_{\mathbf{z}%
}(t_{v})$ by means of the Beran estimator 3 are consistent with the
corresponding event times $t_{1},...,t_{v}$:
\begin{equation}
\mathcal{L}_{\text{Tr1}}=\gamma_{3}\frac{\sum_{i,j}\mathbb{I}\{t_{j}<t_{i}\}
\cdot \sigma(\hat{T}_{i}-\hat{T}_{j})\cdot \delta_{j}}{\sum_{i,j}\mathbb{I}%
\{t_{j}<t_{i}\} \cdot \delta_{j}}.
\end{equation}

The second term $\mathcal{L}_{\text{Tr2}}$ of the loss function $\mathcal{L}%
_{\text{Tr}}$ can be regarded as a regularization for the densities
${\pi(T_{i}|\xi_{\mathbf{z}_{i}}(T_{i}))}$ by using the Beran estimator 3 and
allows us to obtain more elongated trajectories. This can be implemented by
using the likelihood function:
\begin{equation}
\mathcal{L}_{\text{Tr2}}=\gamma_{4}\sum \limits_{i=1}^{n_{u}}\alpha_{i}\log
{\pi(T_{i}|\xi_{\mathbf{z}_{i}}(T_{i})).}%
\end{equation}

Here ${T_{i}}$ are the event times from the training set; $n_{u}$ is the
number of uncensored instances in the training set (only uncensored instances
are used in $\mathcal{L}_{\text{Tr2}}$); ${\xi_{\mathbf{z_{i}}}}$ is the
trajectory for the embedding $\mathbf{x}_{i}$; ${\pi}$ is the density function
computed by using the Beran estimator; $\alpha_{i}$ are smoothing weights
computed as:%

\begin{equation}
\alpha_{i}=\text{softmin}(\{ \pi_{K-M}(T_{1}),\pi_{K-M}(T_{2}),...,\pi
_{K-M}(T_{n_{u}})\})_{i},
\end{equation}
where $\pi_{K-M}(t)$ is the probability density of the event time obtained by
using the Kaplan-Meier estimator over the entire dataset.

Each training epoch includes solving $M$ tasks such that each task consists of
a set of data (\ref{train_epoch}).

The training dataset in this case consists of the following triplets:
$(\mathbf{x}_{i},T_{i},\delta_{i})$, $i=1,...,n$. Thus, $M$ sets of $r+1$
points are selected on each epoch, which are used as the training set, and the
remaining $n-r-1$ points are processed through the model directly and are
passed to the loss function, after which the optimization is performed by the
error backpropagation. After training the model on several epochs, the
background for the Beran estimator is set to the entire training set.

\section{Numerical experiments}

Numerical experiments are performed in the following three directions:

\begin{enumerate}
\item Experiments with synthetic data.

\item Experiments with real data, which illustrate the generation of synthetic
points in accordance with the real dataset.

\item Experiments with real data for constructing the survival regression models.
\end{enumerate}

\subsection{Experiments with synthetic data}

In all experiments, we study the proposed model using instances with two
clusters. The cluster structure of data is used to complicate conditions of
the generation. Instances have two features, i.e., $\mathbf{x}\in
\mathbb{R}^{2}$. They are represented on the graphs in 3D, time is located
along the $Oz$ axis ($\hat{T}$ or $T_{gen}$, to be specified separately). We
perform the unconditional generation. The number of sampled points in
experiments is the same as the number of points in the training dataset. When
performing the conditional generation, we consider both the time generated
using the Gumbel softmax operation and the expected event time.

The following parameters of numerical experiments for synthetic data are used:
the length of embeddings $\mathbf{z}_{i}$ is $8$; the number of embeddings
$\mathbf{z}_{i}$ in the weighting scheme is $m=48$;

Hyperparameters of the loss function (\ref{main_loss}): parameter $\lambda$ in
(\ref{l_wae}) is $40$; $\gamma_{1}=0.5$; $\gamma_{2}=2$; $\gamma_{3}=1$;
$\gamma_{4}=0.05$;

\subsubsection{\textquotedblleft Linear\textquotedblright \ dataset}

\begin{figure}
[ptb]
\begin{center}
\includegraphics[
height=2.8864in,
width=2.8864in
]%
{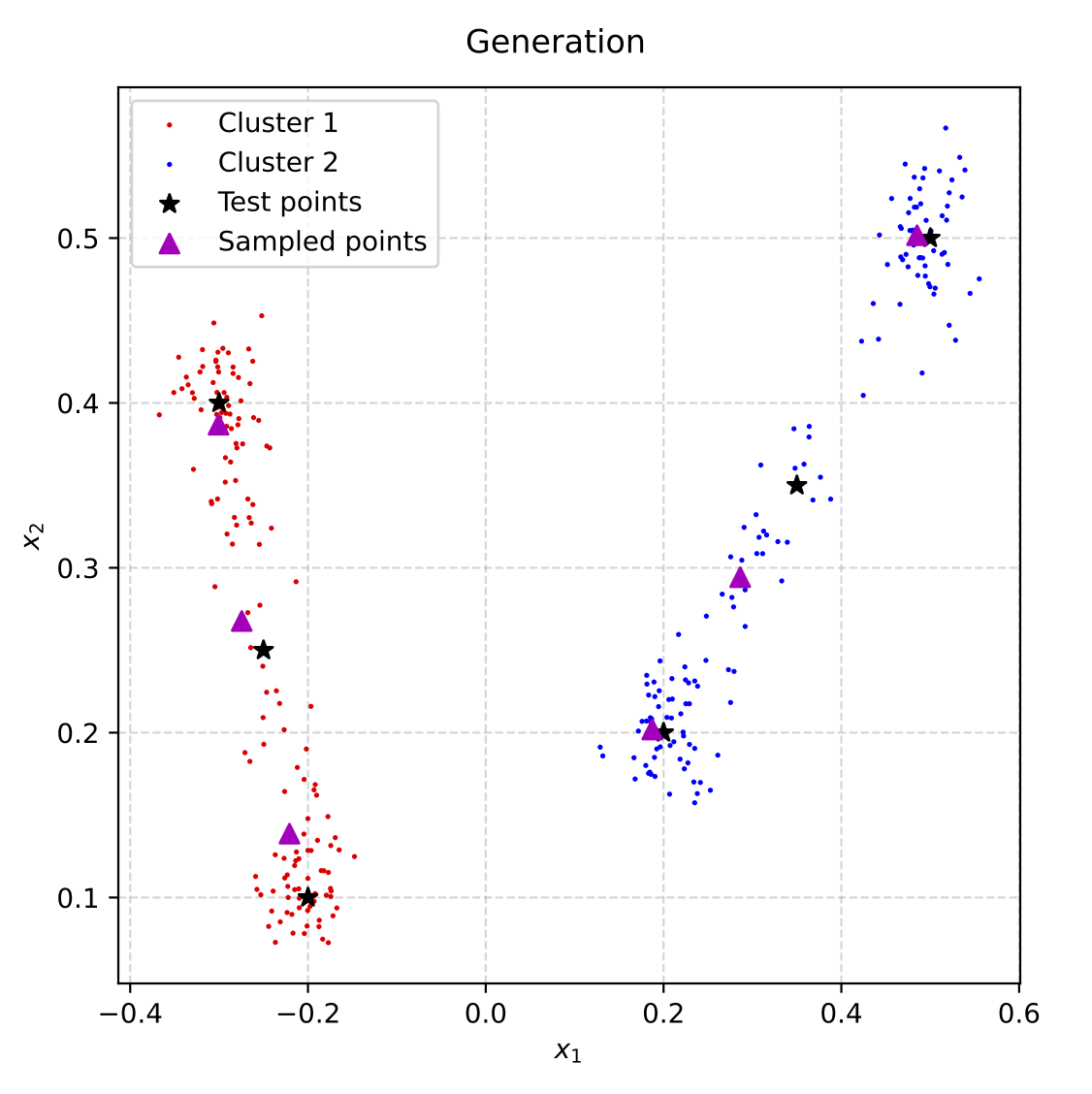}%
\caption{Illustration of generated points $\widehat{\mathbf{x}}$ for the
\textquotedblleft linear\textquotedblright \ dataset}%
\label{f:linear_rec_2d_1}%
\end{center}
\end{figure}

First, we study the linear synthetic dataset which is conditionally called
\textquotedblleft linear\textquotedblright \ because two clusters of feature
vectors are located along straight lines and the event times are uniformly
distributed over each cluster. Clusters are formed by means of four clouds of
normally distributed points. Each point within a cluster is a convex
combination of centers of clouds corresponding to the cluster adding the
normal noise. Coefficients in the convex combination are generated with
respect to the uniform distribution. The obtained clusters are depicted in
Fig. \ref{f:linear_rec_2d_1} in red and blue colors. Fig.
\ref{f:linear_rec_2d_1} illustrates how points $\widehat{\mathbf{x}}$ depicted
by purple triangles are generated for input points $\mathbf{x}$ depicted by stars.%

\begin{figure}
[ptb]
\begin{center}
\includegraphics[
height=3.1819in,
width=3.1819in
]%
{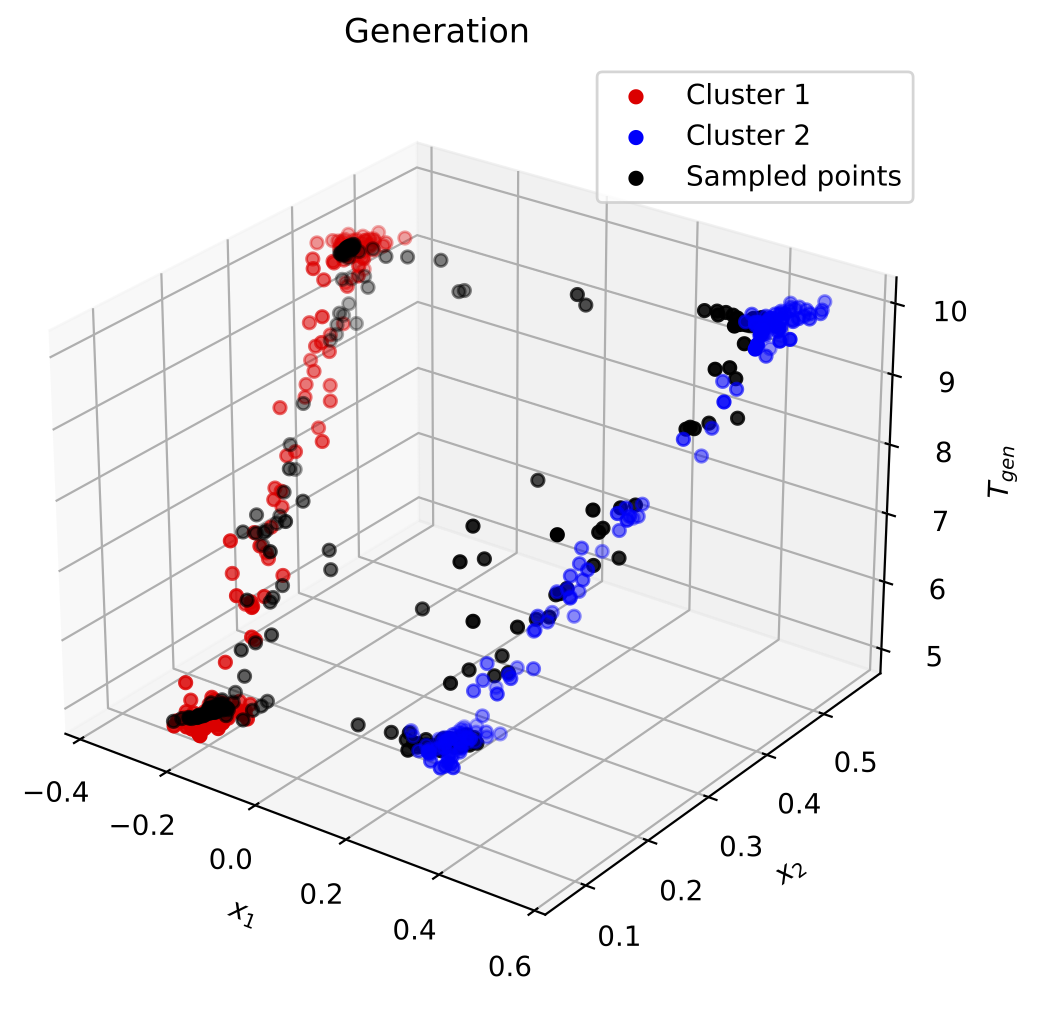}%
\caption{Generated points $(\widehat{\mathbf{x}},T_{gen})$ for the
\textquotedblleft linear\textquotedblright \ dataset}%
\label{f:linear_rec_t_gen}%
\end{center}
\end{figure}

Fig. \ref{f:linear_rec_t_gen} illustrates the same generation of points
$\widehat{\mathbf{x}}$ depicted by black markers jointly with generated times
to event $T_{gen}$. One can see from Fig. \ref{f:linear_rec_t_gen} that most
points are close to the dataset points from the corresponding clusters.
However, there are a few points located between clusters, which are generated
incorrectly. Figs. \ref{f:linear_rec_2d_1} and \ref{f:linear_rec_t_gen} show
that the proposed model mainly correctly reconstructs feature vectors and
correctly generates the event times.%

\begin{figure}
[ptb]
\begin{center}
\includegraphics[
height=2.6164in,
width=5.1976in
]%
{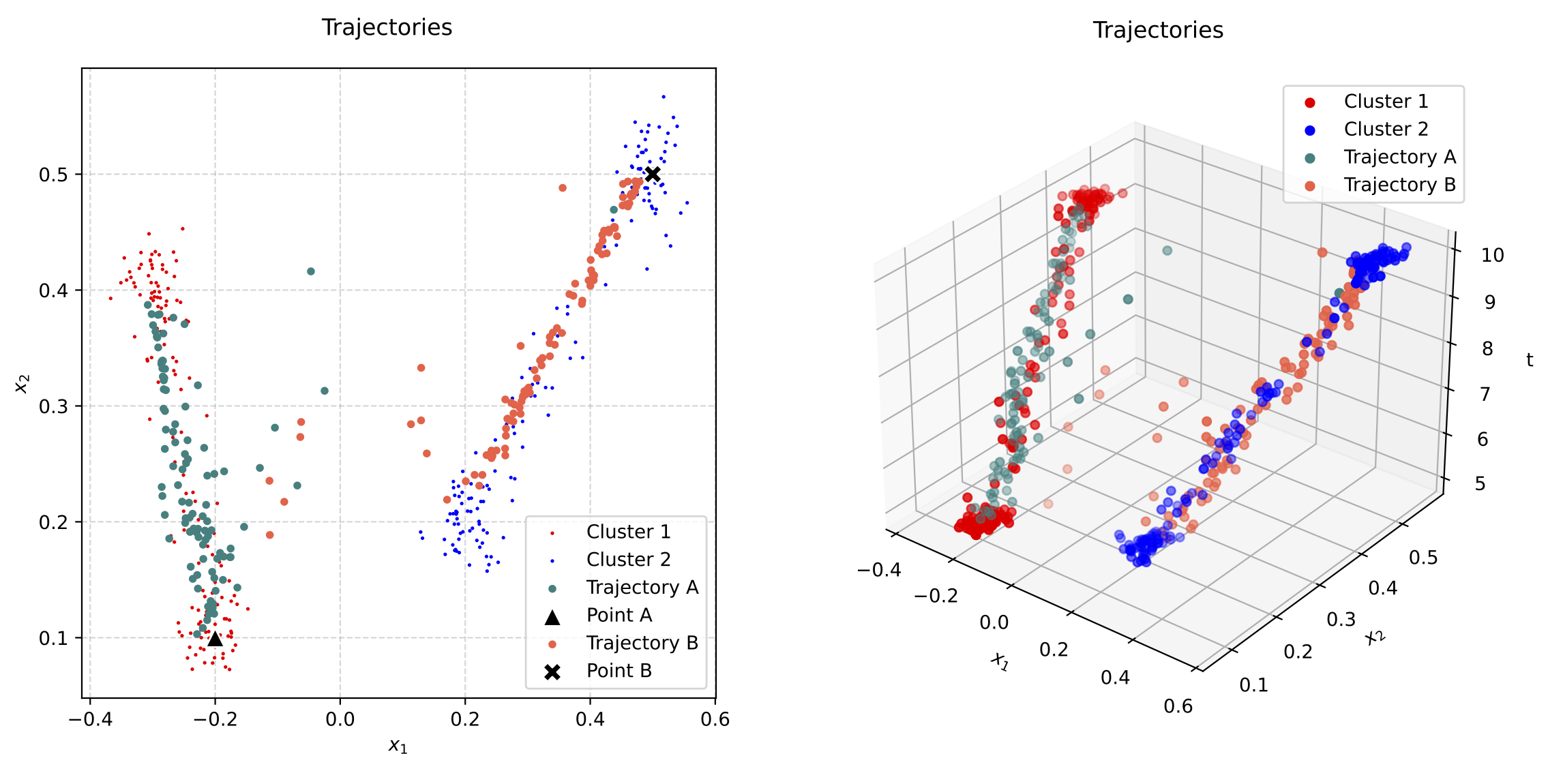}%
\caption{Generated trajectories for the \textquotedblleft
linear\textquotedblright \ dataset}%
\label{f:linear_traj_2d_3d}%
\end{center}
\end{figure}

\begin{figure}
[ptb]
\begin{center}
\includegraphics[
height=2.7395in,
width=2.9049in
]%
{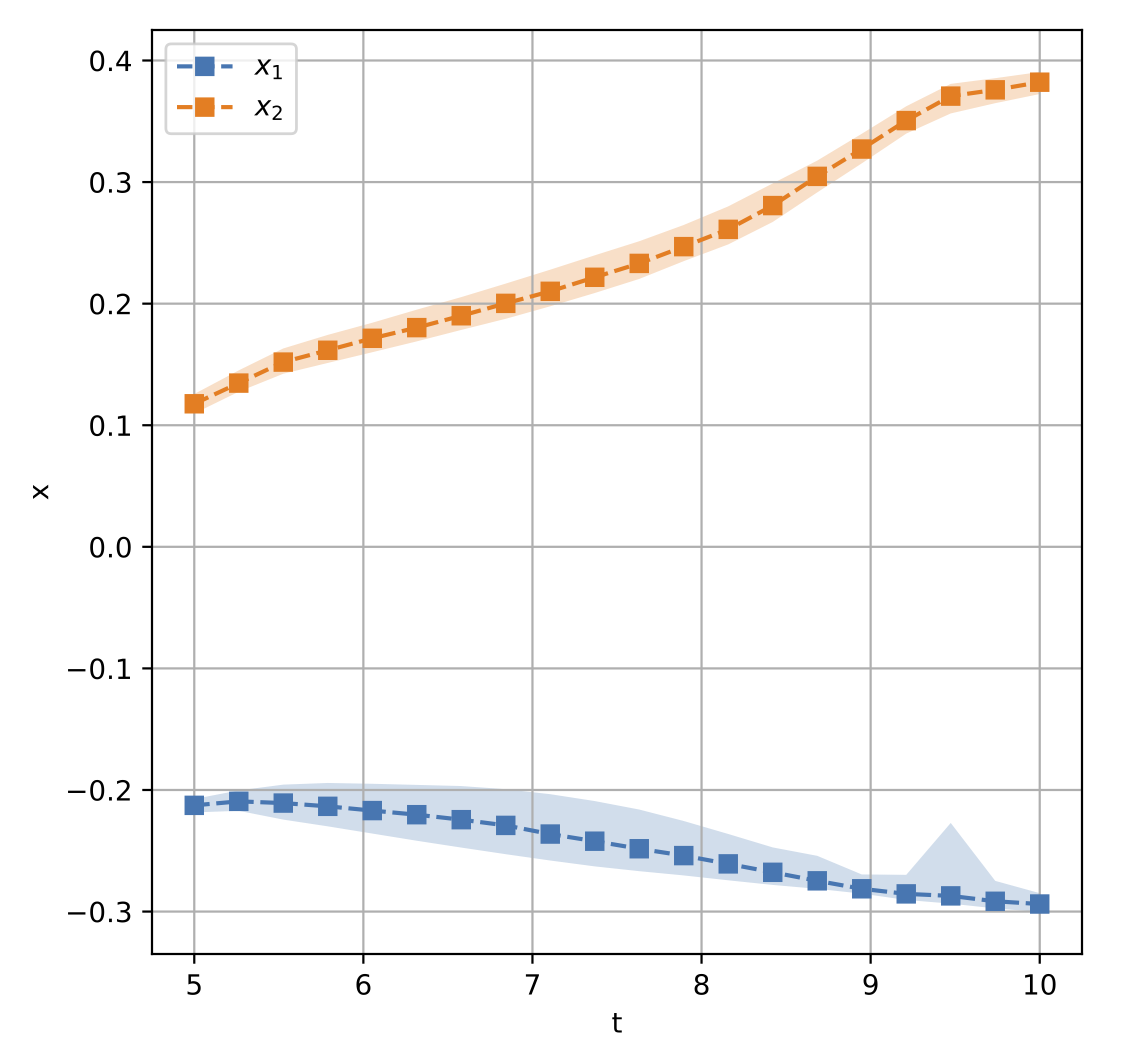}%
\caption{Generated trajectories of each feature as functions of the time for
the \textquotedblleft linear\textquotedblright \ dataset}%
\label{f:linear_traj_sep}%
\end{center}
\end{figure}

Generated trajectories for points A and B from the first and the second
clusters, respectively, of the \textquotedblleft linear\textquotedblright%
\ dataset are illustrated in Fig. \ref{f:linear_traj_2d_3d} where the left
picture shows only the generated points $\xi_{\mathbf{x}}$ without time
moments, the right picture shows points of the same trajectory taking into
account time moments $t_{1},...,t_{v}$. It can be seen from Fig.
\ref{f:linear_traj_2d_3d} that the generated trajectory corresponds to the
location of points from the dataset. An example of the generated feature
trajectories as functions of the time for the \textquotedblleft
linear\textquotedblright \ dataset also shown in Fig. \ref{f:linear_traj_sep}.
The point A is taken to generate the trajectory. It is important to note that
the trajectories of each feature are rather smooth. This is due to the
weighting procedure which is used to generate $\xi_{\mathbf{z}}(t)$ in the
embedding space and due to the correct reconstruction of points by the VAE.%

%

\subsubsection{Two parabolas}

Let us consider an illustrative example with a dataset which is similar to the
well-known \textquotedblleft two moons\textquotedblright \ dataset, which can
be found in the Python Scikit-learn package. In contrast to the use of the
original \textquotedblleft two moons\textquotedblright \ dataset, we complicate
the task by considering two different clusters (parabolas) of data, but with
similar event times. The event times are generated linearly from the feature
$x_{1}$, so the values on each branch of each parabola are symmetrically located.

\begin{figure}
[ptb]
\begin{center}
\includegraphics[
height=2.6243in,
width=5.0437in
]%
{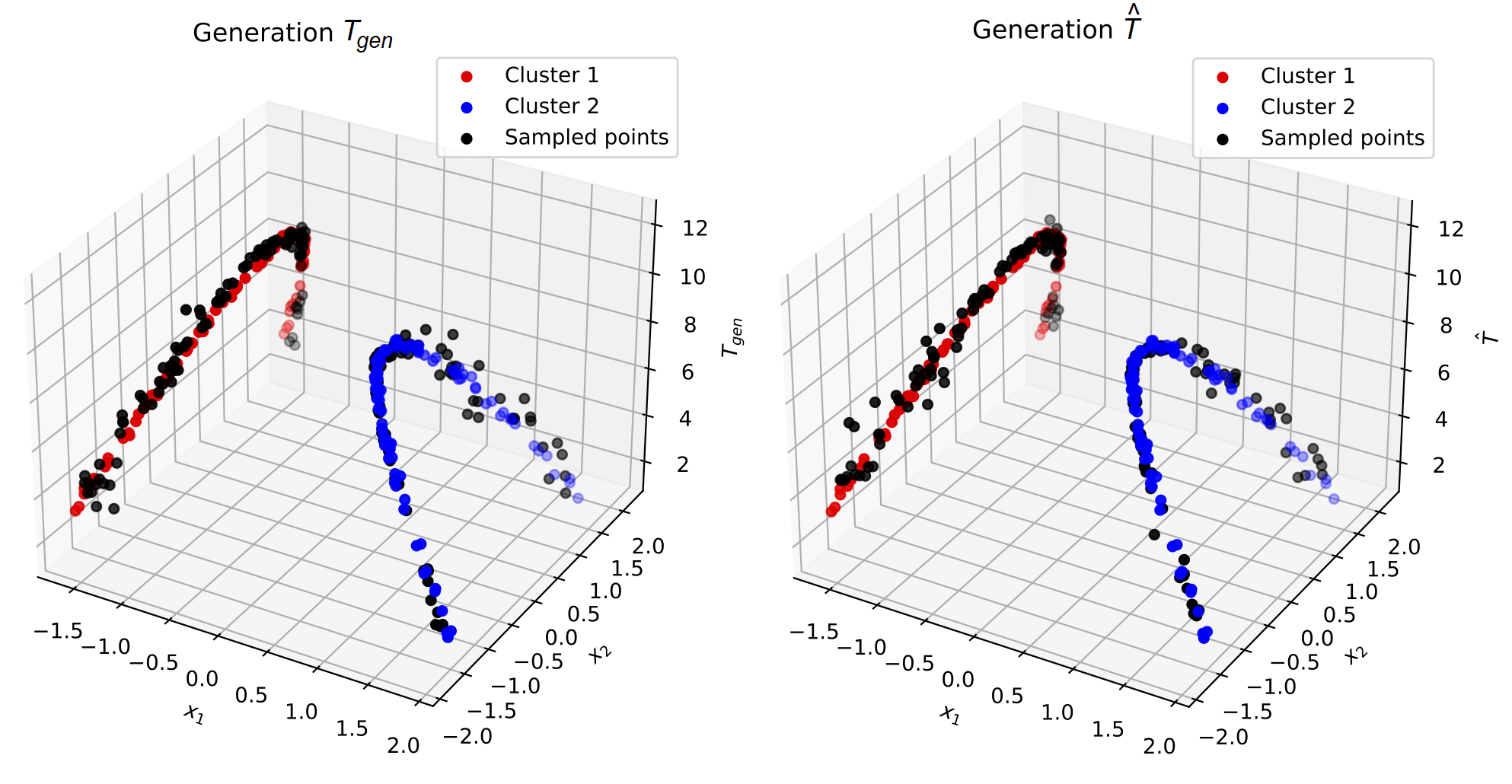}%
\caption{Generation of $T_{gen}$ and $\widehat{T}$ for the \textquotedblleft
two parabolas\textquotedblright \ dataset }%
\label{f:moons_rec_t_gen_e}%
\end{center}
\end{figure}

Results of generation of $T_{gen}$ and $\widehat{T}$ are depicted in the left
and the right pictures of Fig. \ref{f:moons_rec_t_gen_e}. One can see from
Fig. \ref{f:moons_rec_t_gen_e} that points are mainly correctly generated.
Moreover, the expected event time $\widehat{T}$ has smaller fluctuations than
the generated one $T_{gen}$. This fact demonstrates that the model properly generates.%

\begin{figure}
[ptb]
\begin{center}
\includegraphics[
height=2.6384in,
width=5.0262in
]%
{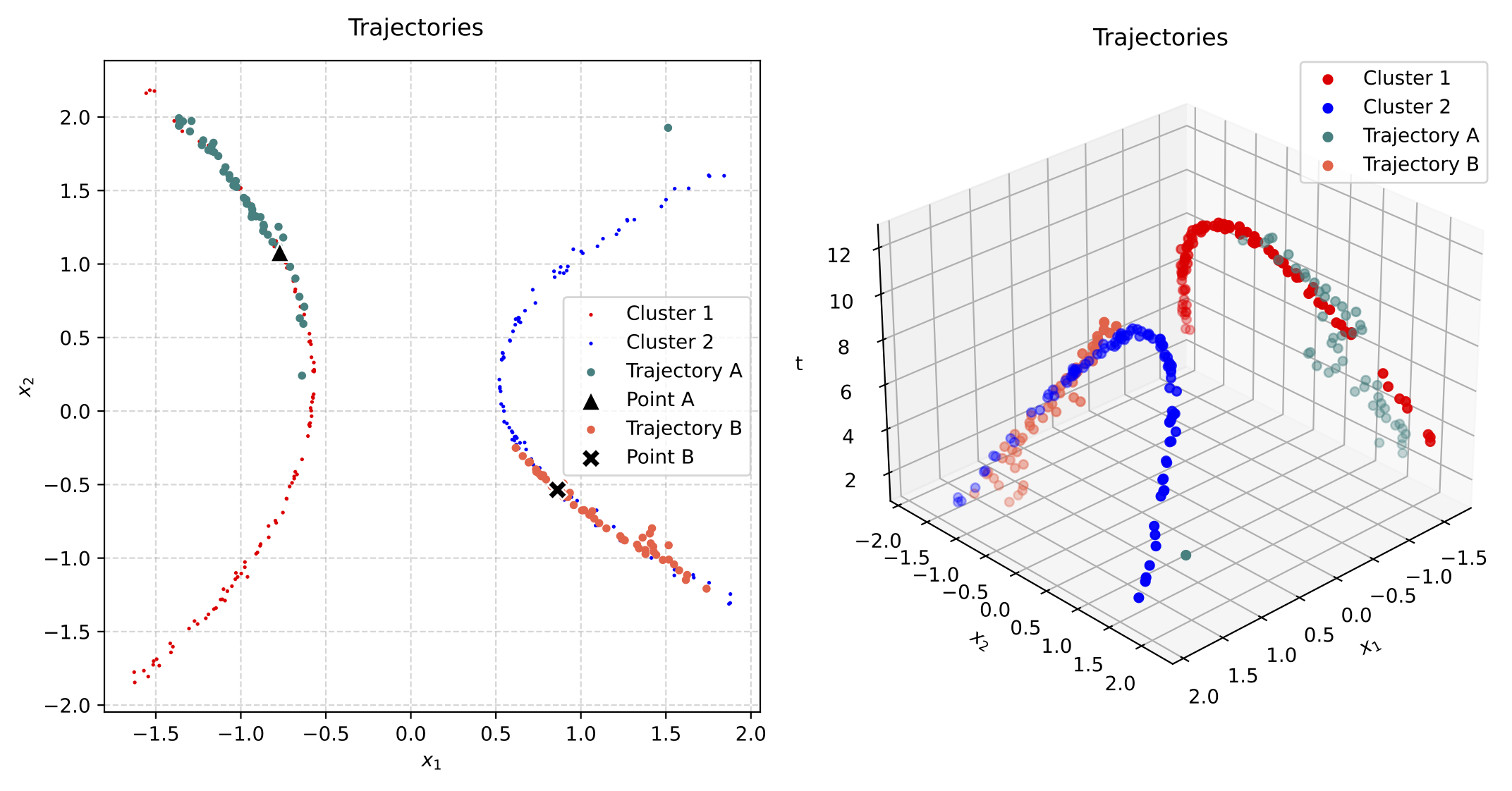}%
\caption{Generation of trajectories for points A and B by using the
\textquotedblleft two parabolas\textquotedblright \ dataset }%
\label{f:moons_traj_2d_3d}%
\end{center}
\end{figure}

Fig. \ref{f:moons_traj_2d_3d} illustrates how trajectories for points A and B
are generated. The left and the right pictures in Fig.
\ref{f:moons_traj_2d_3d} show the trajectories without the event times and
with the times. It is explicitly seen that trajectories are generated on
branches of the parabolas.%

\begin{figure}
[ptb]
\begin{center}
\includegraphics[
height=2.7026in,
width=2.8046in
]%
{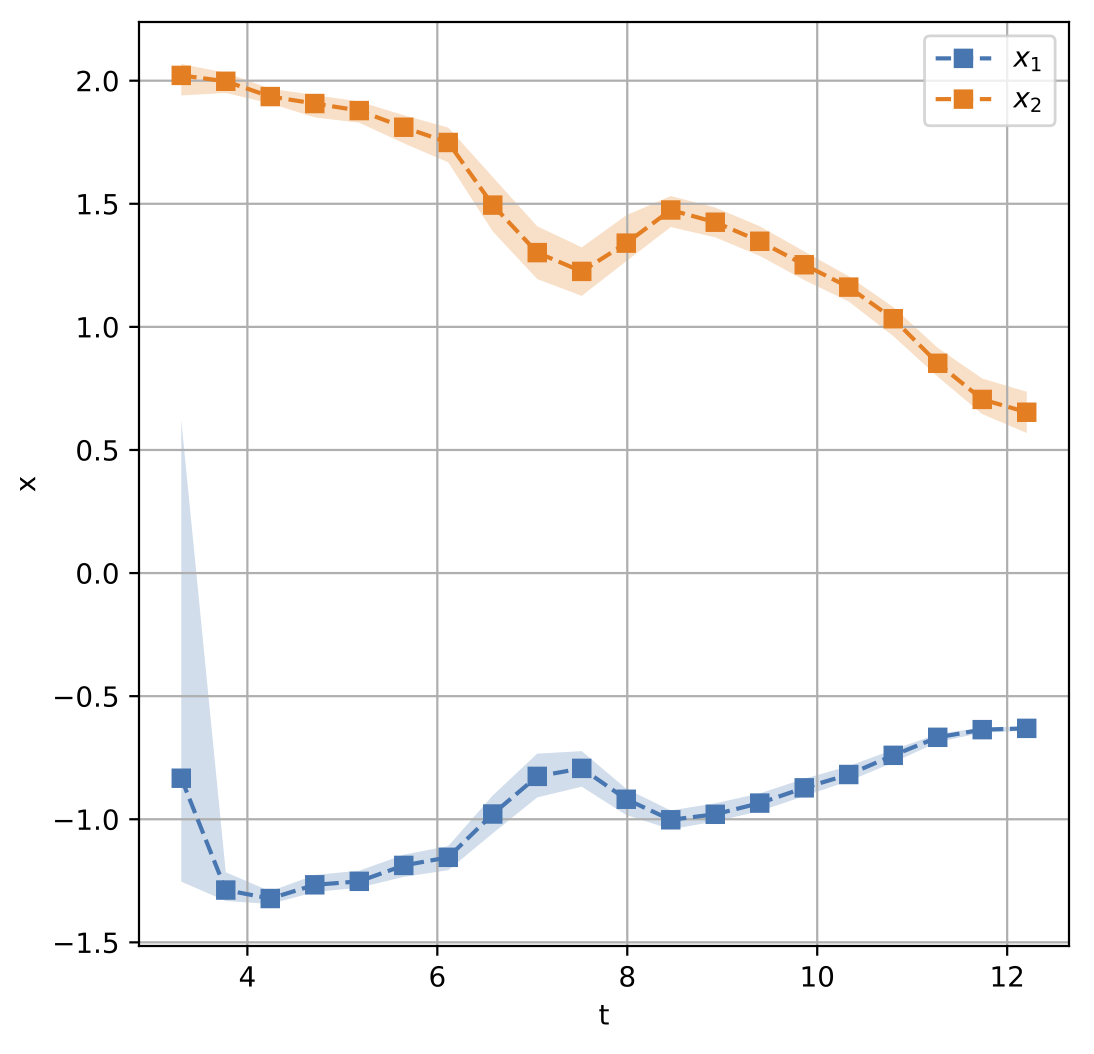}%
\caption{Generated trajectories of each feature as functions of the time for
the \textquotedblleft two parabolas\textquotedblright \ dataset }%
\label{f:moons_traj_sep}%
\end{center}
\end{figure}

Generated feature trajectories as functions of the time for the
\textquotedblleft two parabolas\textquotedblright \ dataset are shown in Fig.
\ref{f:moons_traj_sep}. The point A is taken to generate the trajectory.%

\subsubsection{Two circles}

\begin{figure}
[ptb]
\begin{center}
\includegraphics[
height=3.0122in,
width=3.2153in
]%
{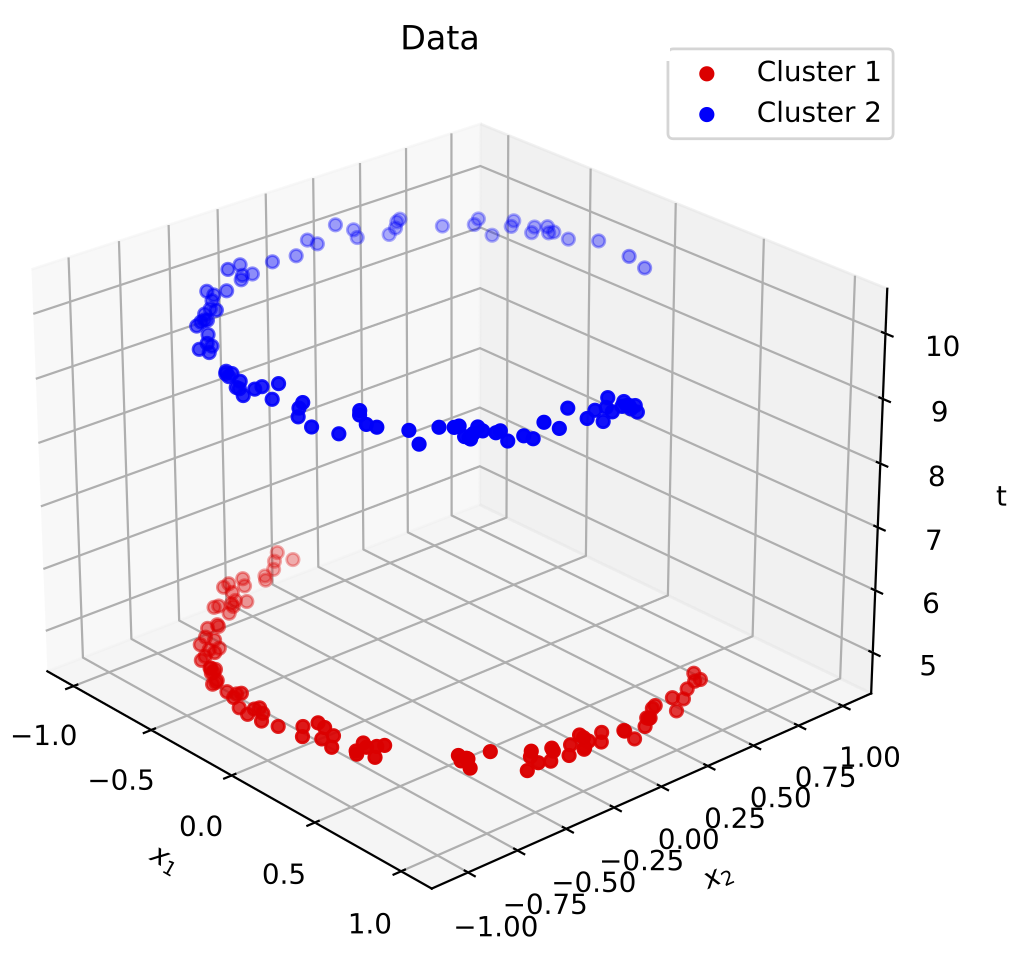}%
\caption{The \textquotedblleft two circles\textquotedblright \ dataset}%
\label{f:two_circles}%
\end{center}
\end{figure}

Another interesting synthetic dataset consists of two circles as it is shown
in Fig. \ref{f:two_circles}. More precisely, we are conducting the experiment
not with full-fledged circles, but with their sectors. The essence of the
experiment is that there are regions where the event time seriously differs
for very close feature vectors. At the same time, the event times for points
belonging to each circle are slightly differ. They are generated with a small
noise.%

\begin{figure}
[ptb]
\begin{center}
\includegraphics[
height=2.5206in,
width=5.4228in
]%
{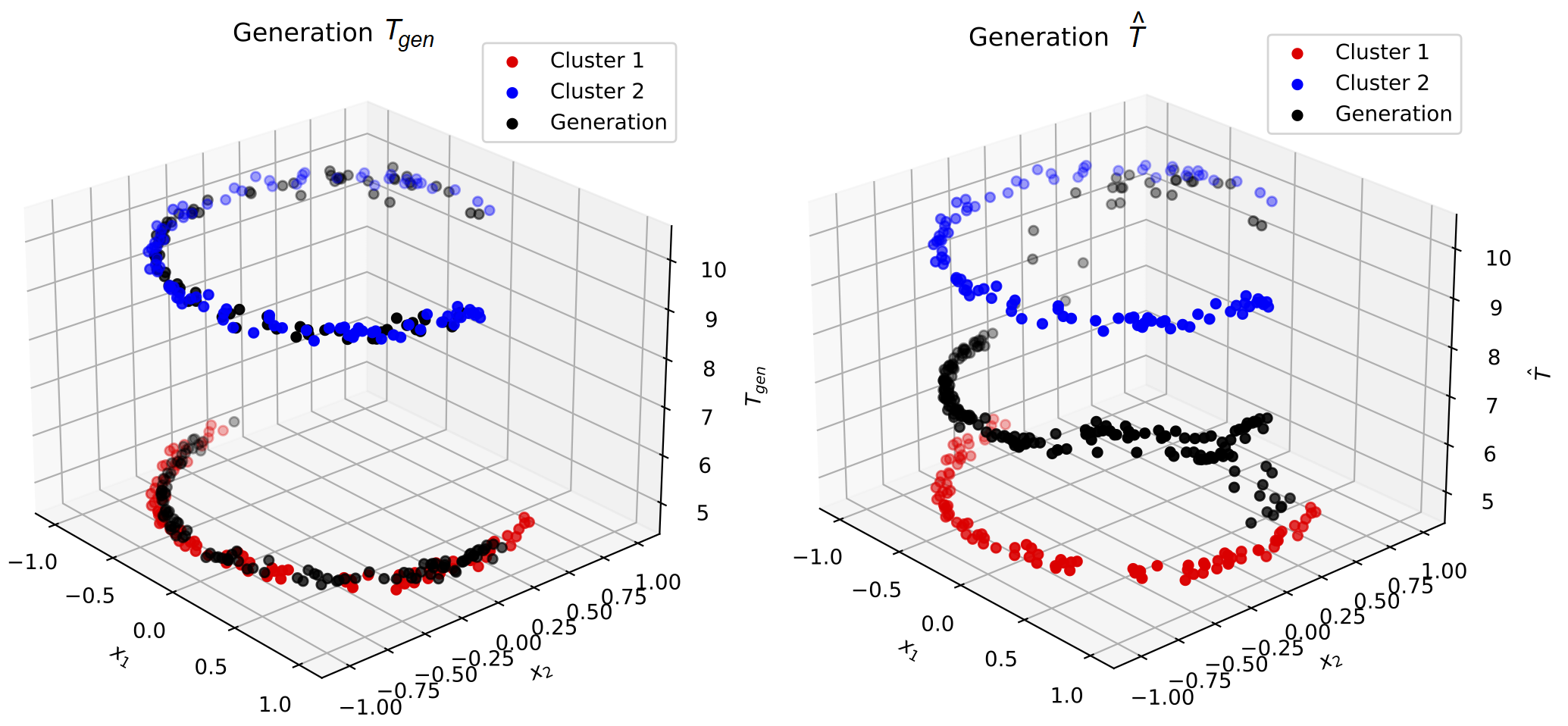}%
\caption{Generation of $T_{gen}$ and $\widehat{T}$ for the \textquotedblleft
two circles\textquotedblright \ dataset }%
\label{f:two_circles_t_gen_e}%
\end{center}
\end{figure}

The left and the right pictures in Fig. \ref{f:two_circles_t_gen_e} show
results of generation of the event times $T_{gen}$ and the expected times
$\widehat{T}$. It can be seen from Fig. \ref{f:two_circles_t_gen_e} that the
the event times $T_{gen}$ are correctly generated. This is due to the use of
the Gumbel softmax operation. However, it follows from the right picture in
Fig. \ref{f:two_circles_t_gen_e} that the expected times give a strong bias
which is caused by the multimodality of the probability distribution of the variables.

The task of the trajectory generation is not studied here because trajectories
are simply be vertical in the 3D pictures in the overlapped area.%

\subsection{Experiments with real data}

The well-known real datasets, including the Veteran dataset, the WHAS500
dataset, and the GBSG2 dataset, are used for numerical experiments.

\subsubsection{Veteran dataset}

The first dataset is the \emph{Veterans' Administration Lung Cancer Study
(Veteran) Dataset} \cite{Kalbfleisch-Prentice-1980} which contains data on 137
males with advanced inoperable lung cancer. The subjects were randomly
assigned to either a standard chemotherapy treatment or a test chemotherapy
treatment. The dataset can be obtained via the \textquotedblleft
survival\textquotedblright \ R package or the Python \textquotedblleft
scikit-survival\textquotedblright \ package.

\begin{figure}
[ptb]
\begin{center}
\includegraphics[
height=2.3557in,
width=3.1375in
]%
{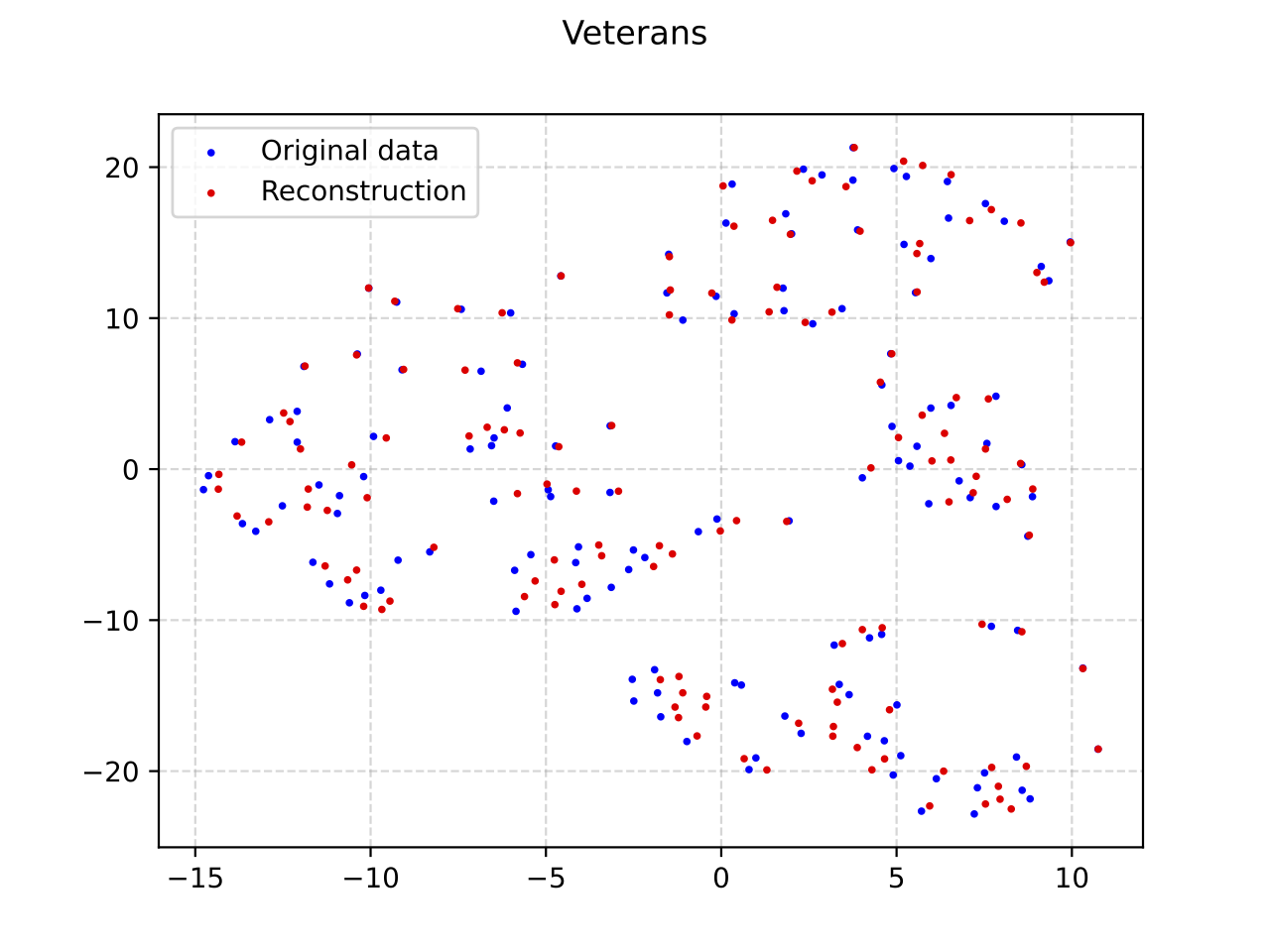}%
\caption{Original and generated instances for the Veteran dataset}%
\label{f:veterans_rec}%
\end{center}
\end{figure}

\begin{figure}
[ptb]
\begin{center}
\includegraphics[
height=2.3947in,
width=3.1894in
]%
{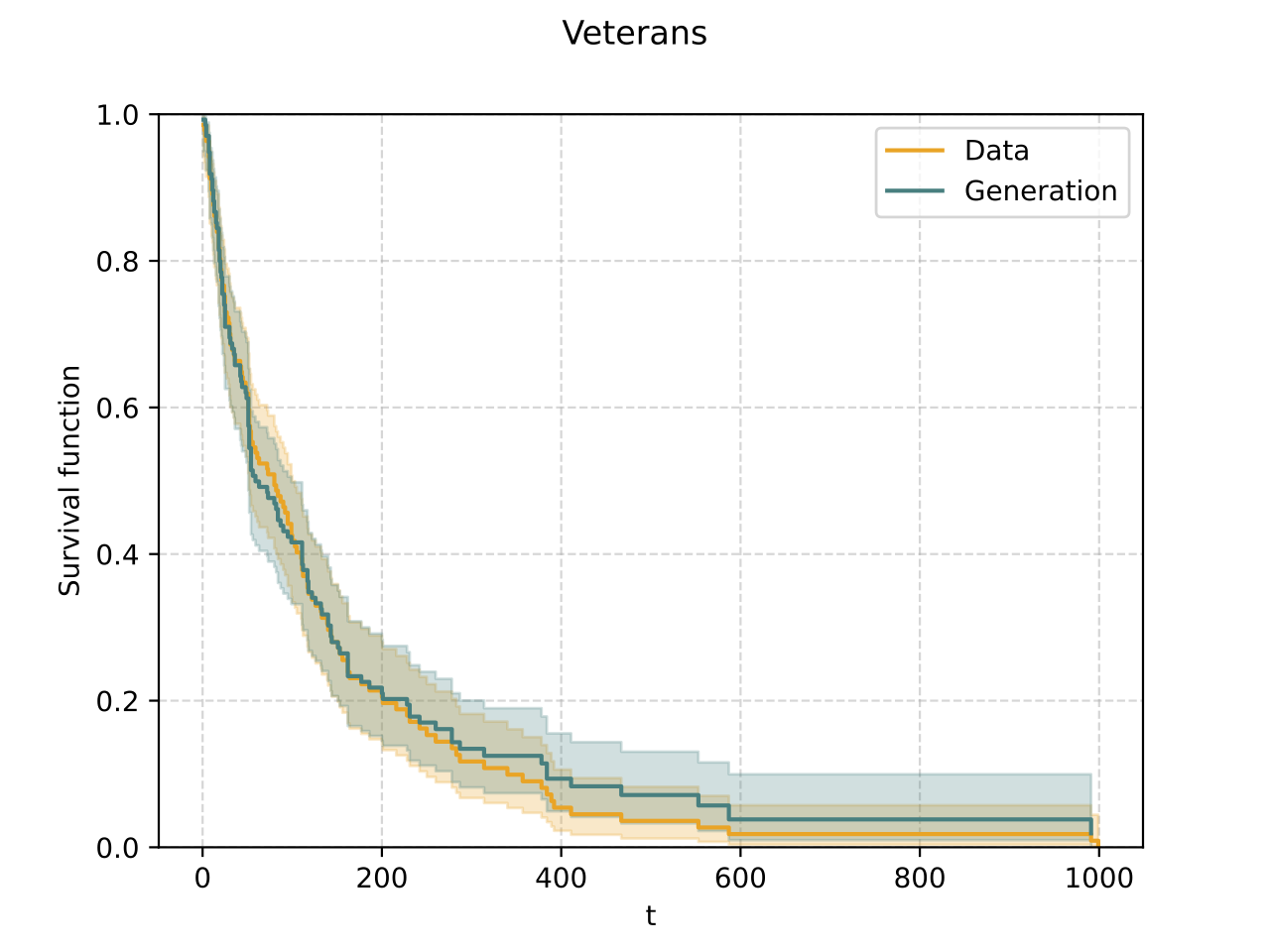}%
\caption{SFs constructed by means of the Kaplan-Meier estimator for original
and generated data for the Veteran dataset}%
\label{f:veterans_km}%
\end{center}
\end{figure}

By training and using the proposed model, we generate new instances in
accordance with the Veteran dataset. Results are depicted in Fig.
\ref{f:veterans_rec} where original points and the reconstructed points are
shown in blue and red colors, respectively. The t-SNE method
\cite{van2008visualizing} is used to depict points in the 2D plot. It can be
seen from Fig. \ref{f:veterans_rec} that the generated points support the
complex cluster structure of the dataset. In order to study how the generated
instances are close to the original data, we compute SFs for these two sets of
instances by using the Kaplan-Meier estimator. The SFs are shown in Fig.
\ref{f:veterans_km}. It can be seen from Fig. \ref{f:veterans_km} that the SFs
are very close to each other. This implies that the model provides a proper generation.

\begin{figure}
[ptb]
\begin{center}
\includegraphics[
height=3.1825in,
width=3.1825in
]%
{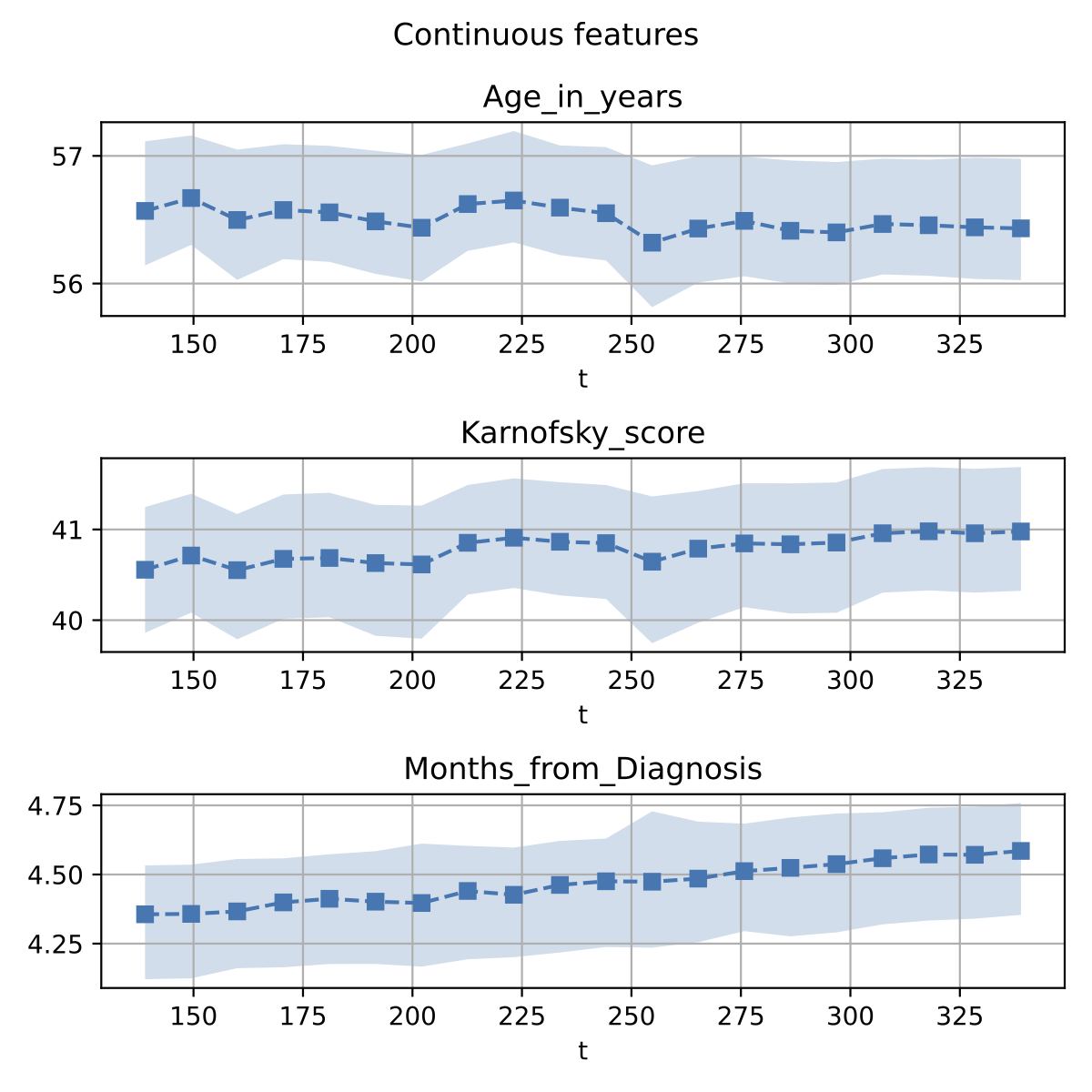}%
\caption{Trajectories of the continuous features generated for the Veteran
dataset}%
\label{f:veterans_continuous_3}%
\end{center}
\end{figure}

In order to depict the generated trajectory, we show how separate features
should be changed to achieve a certain event time. The corresponding
trajectories of the continuous features are depicted in Fig.
\ref{f:veterans_continuous_3}. It is obvious that the feature
\textquotedblleft Age in years\textquotedblright \ cannot be changed in real
life. However, our aim is to show that the trajectory is correctly generated.
We can see that the age should be decreased to increase the event time. This
indicates that the model correctly generates the trajectory.%

%

%

\subsubsection{WHAS500 dataset}

Another dataset is the \emph{Worcester Heart Attack Study (WHAS500) Dataset}
\cite{Hosmer-Lemeshow-May-2008}. It contains data on 500 patients having 14
features. The endpoint is death, which occurred for 215 patients (43.0\%). The
dataset can be obtained via the \textquotedblleft smoothHR\textquotedblright%
\ R package or the Python \textquotedblleft scikit-survival\textquotedblright \ package.

\begin{figure}
[ptb]
\begin{center}
\includegraphics[
height=2.4275in,
width=3.2327in
]%
{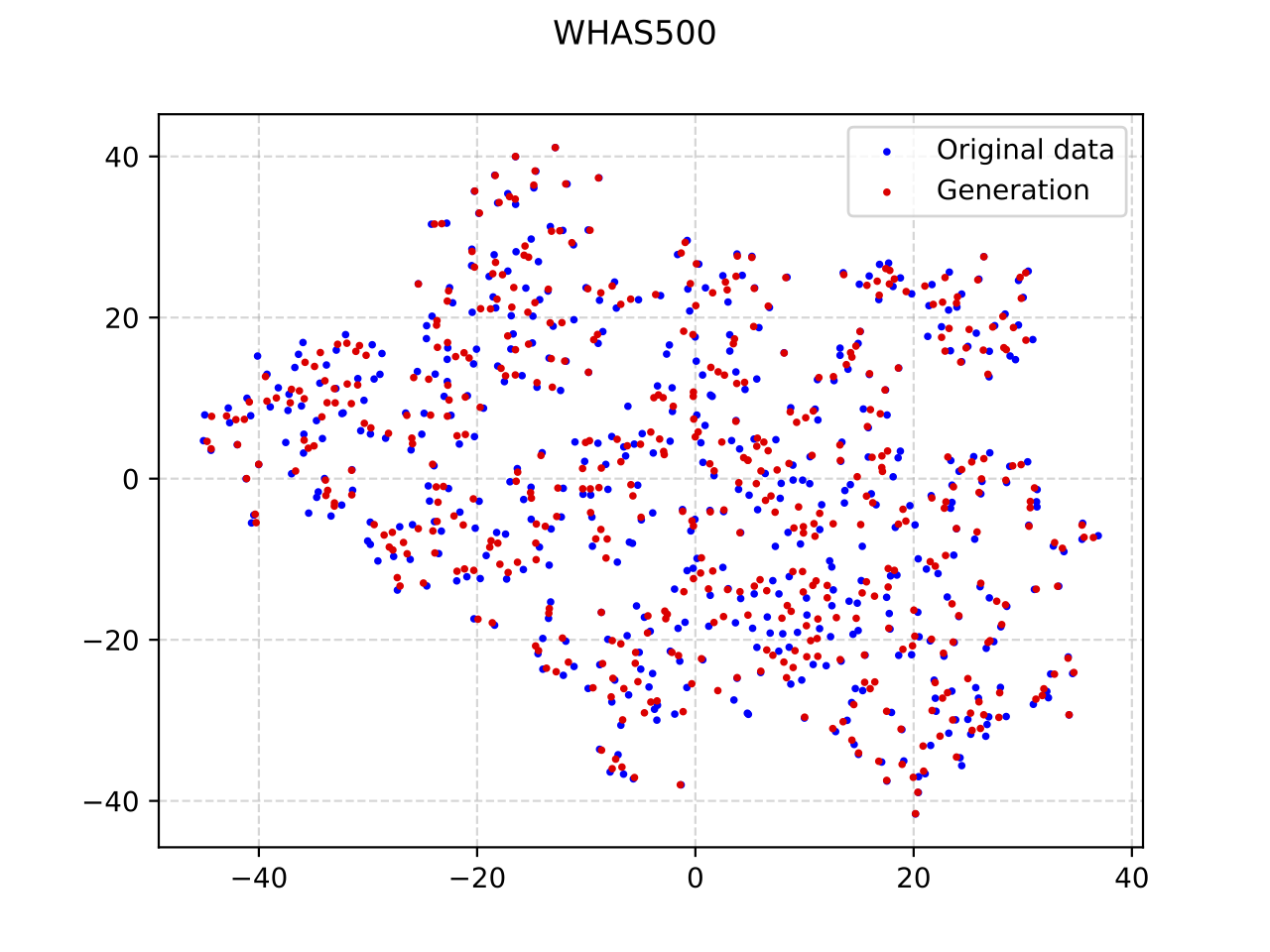}%
\caption{Original and generated instances for the WHAS500 dataset}%
\label{f:whas500_rec}%
\end{center}
\end{figure}

\begin{figure}
[ptb]
\begin{center}
\includegraphics[
height=2.4267in,
width=3.2327in
]%
{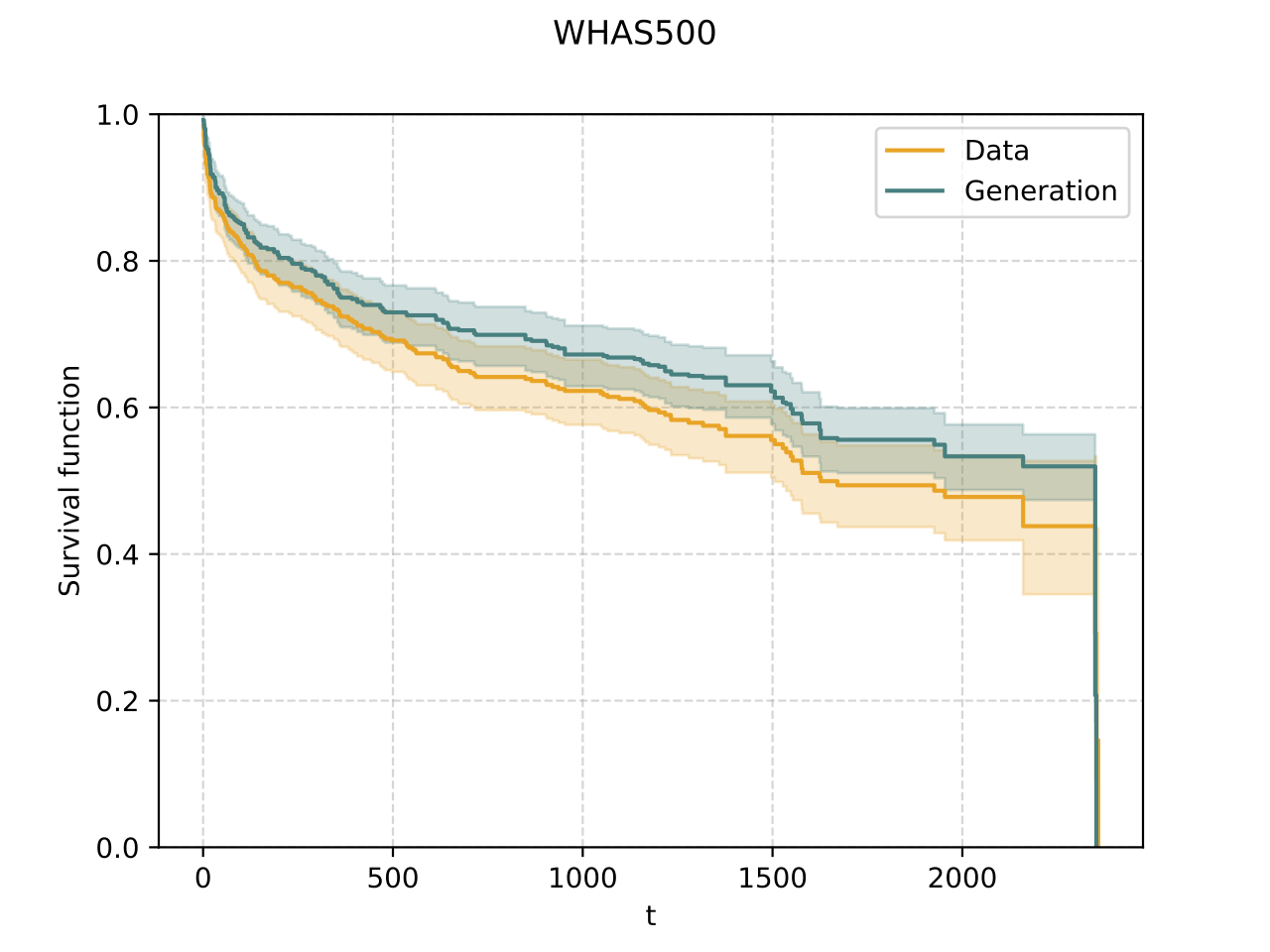}%
\caption{SFs constructed by means of the Kaplan-Meier estimator for original
and generated data for the WHAS500 dataset}%
\label{f:whas500_km}%
\end{center}
\end{figure}

\begin{figure}
[ptb]
\begin{center}
\includegraphics[
height=2.7562in,
width=4.5602in
]%
{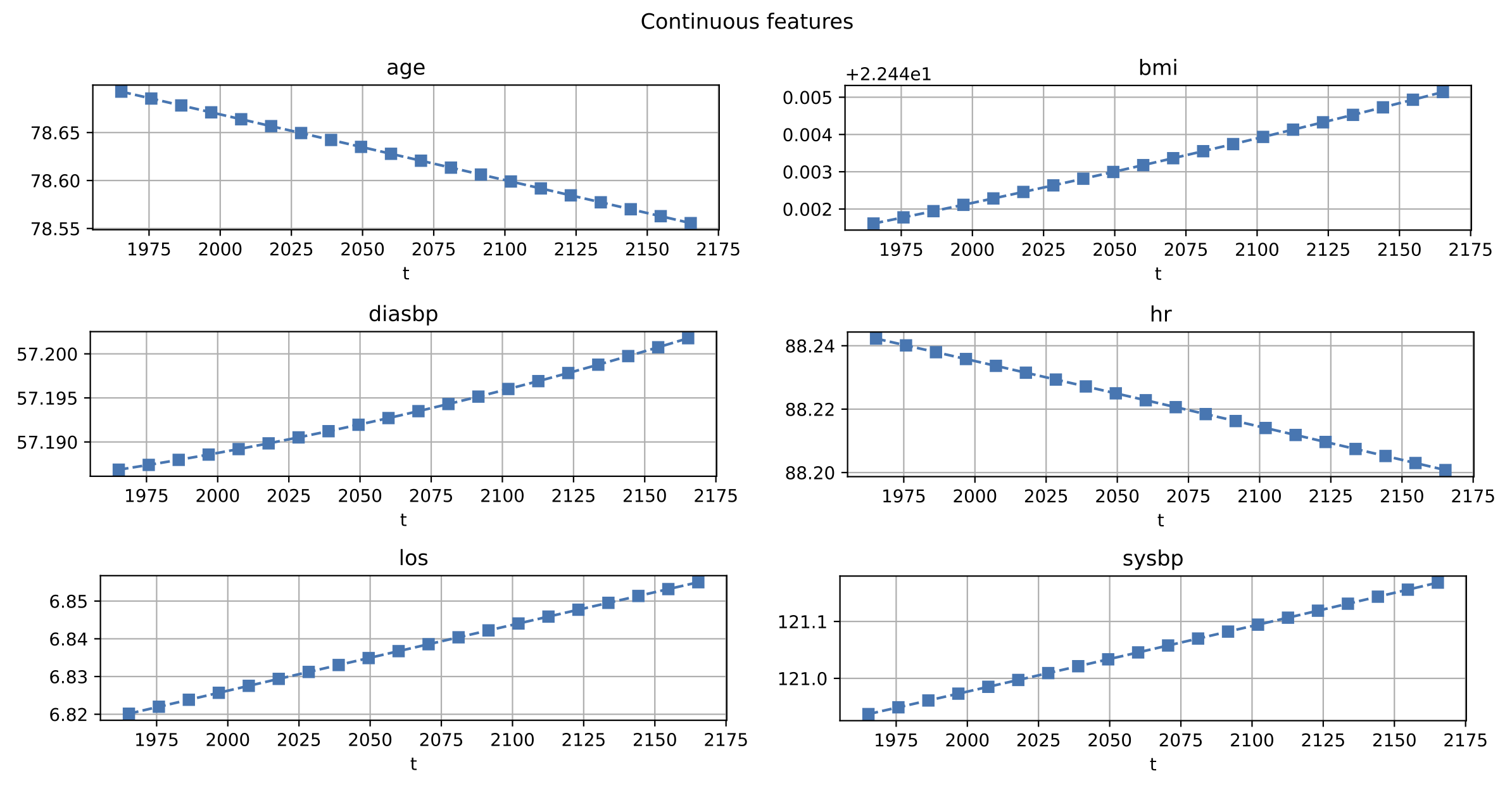}%
\caption{Trajectories of the continuous features generated for the WHAS500
dataset}%
\label{f:whas500_continuous_2}%
\end{center}
\end{figure}

Similar results of experiments are shown in Figs. \ref{f:whas500_rec}%
-\ref{f:whas500_continuous_2}. In particular, original and generated points
are depicted in Fig. \ref{f:whas500_rec} by using the t-SNE method in blue and
red colors, respectively. SFs for original and generated points by using the
Kaplan-Meier estimator are shown in Fig. \ref{f:whas500_km}. Trajectories of
the continuous features are depicted in Fig. \ref{f:whas500_continuous_2}.%

%

%

\subsubsection{GBSG2 dataset}

The next dataset is the \emph{German Breast Cancer Study Group 2 (GBSG2)
Dataset} \cite{Sauerbrei-Royston-1999} which contains observations of 686
women. Every instance is characterized by 10 features, including age of the
patients in years, menopausal status, tumor size, tumor grade, number of
positive nodes, hormonal therapy, progesterone receptor, estrogen receptor,
recurrence free survival time, censoring indicator (0 - censored, 1 - event).
The dataset can be obtained via the \textquotedblleft
TH.data\textquotedblright \ R package or the Python \textquotedblleft
scikit-survival\textquotedblright \ package.

\begin{figure}
[ptb]
\begin{center}
\includegraphics[
height=2.4734in,
width=3.2932in
]%
{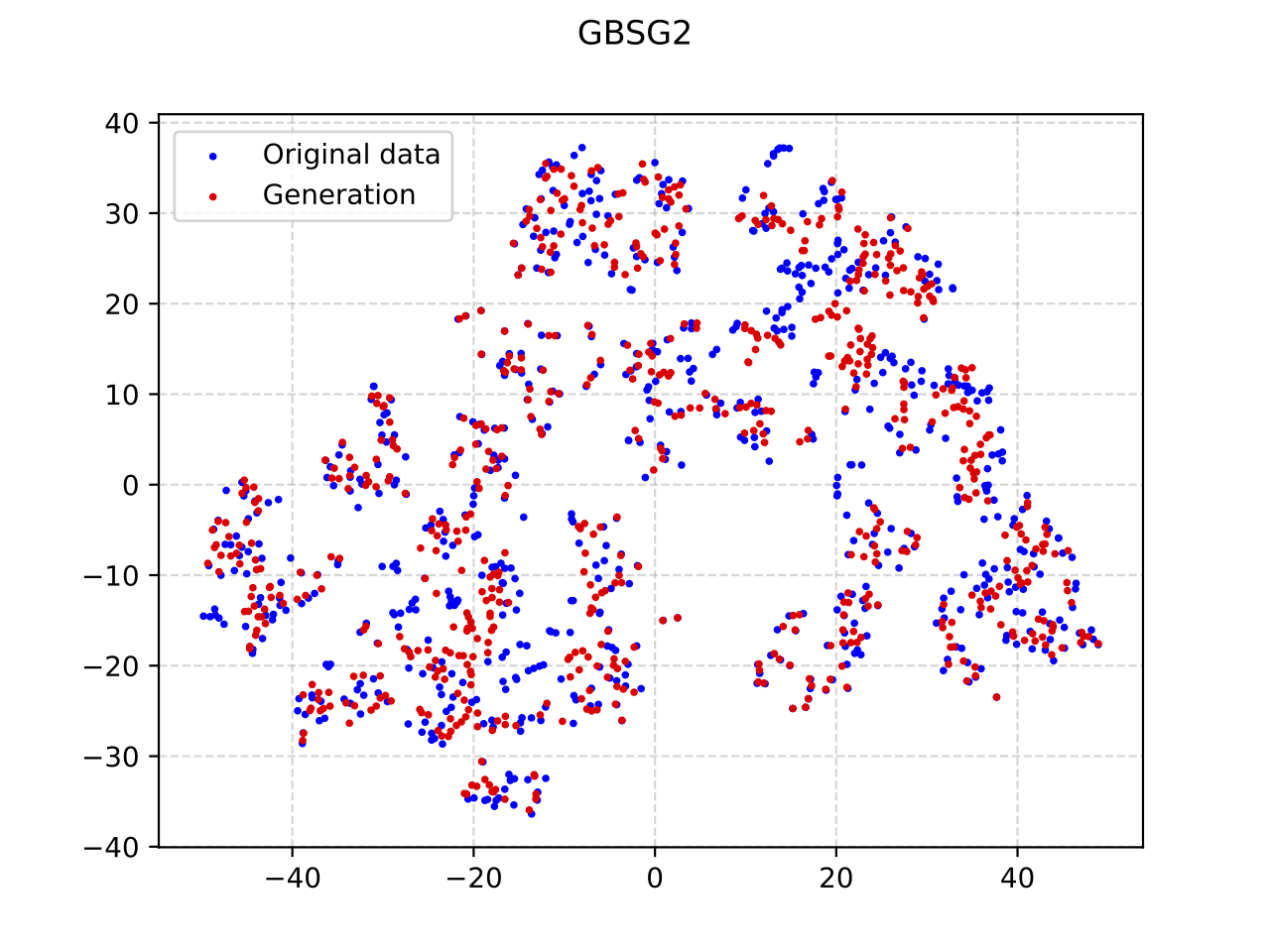}%
\caption{Original and generated instances for the GBSG2 dataset}%
\label{f:gbsg2_rec}%
\end{center}
\end{figure}

\begin{figure}
[ptb]
\begin{center}
\includegraphics[
height=2.4993in,
width=3.3278in
]%
{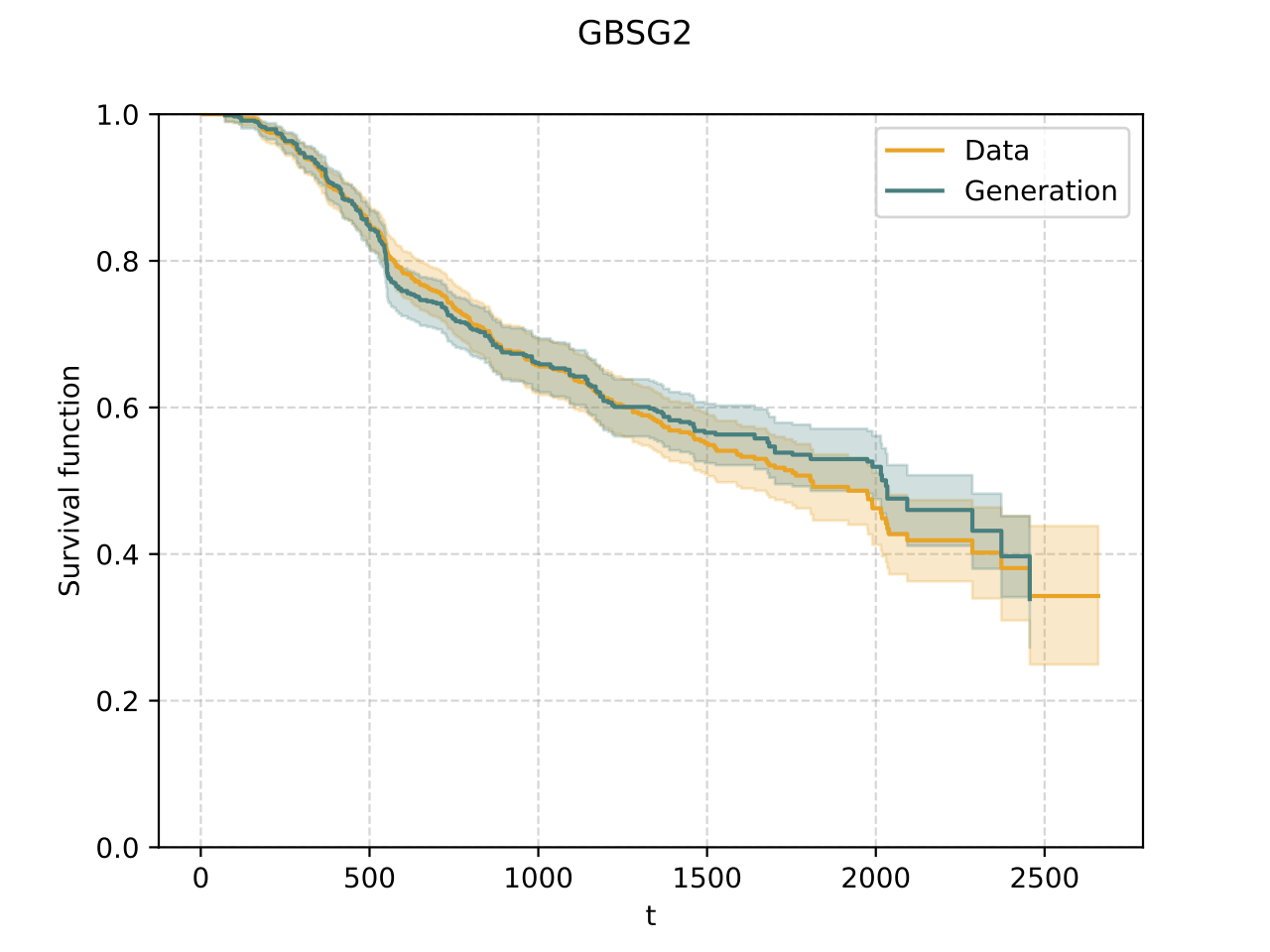}%
\caption{SFs constructed by means of the Kaplan-Meier estimator for original
and generated data for the GBSG2 dataset}%
\label{f:gbsg2_km}%
\end{center}
\end{figure}

\begin{figure}
[ptb]
\begin{center}
\includegraphics[
height=2.3722in,
width=4.2566in
]%
{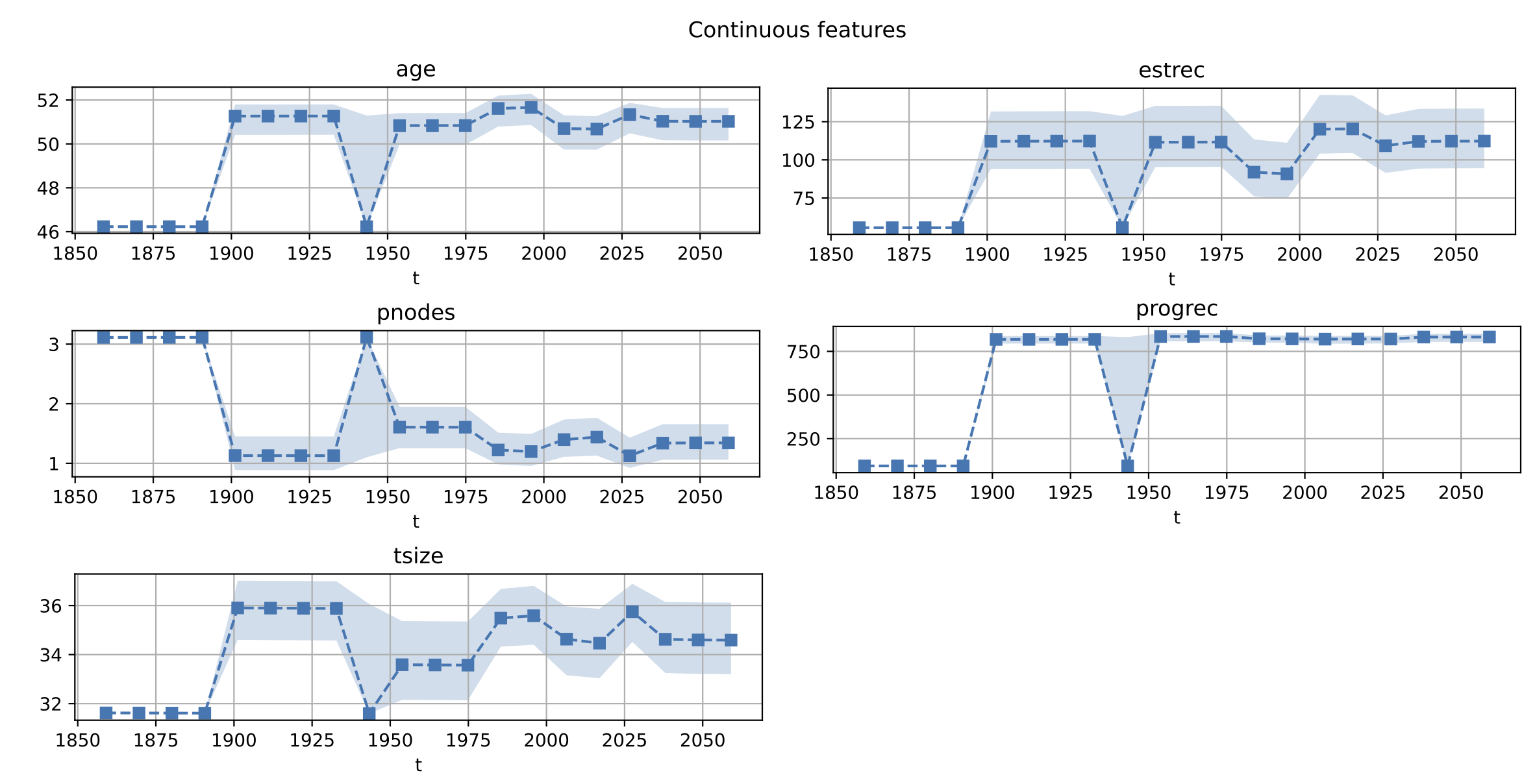}%
\caption{Trajectories of the continuous features generated for the GBSG2
dataset}%
\label{f:gbsg2_continuous_1}%
\end{center}
\end{figure}

\begin{figure}
[ptb]
\begin{center}
\includegraphics[
height=2.4267in,
width=2.4267in
]%
{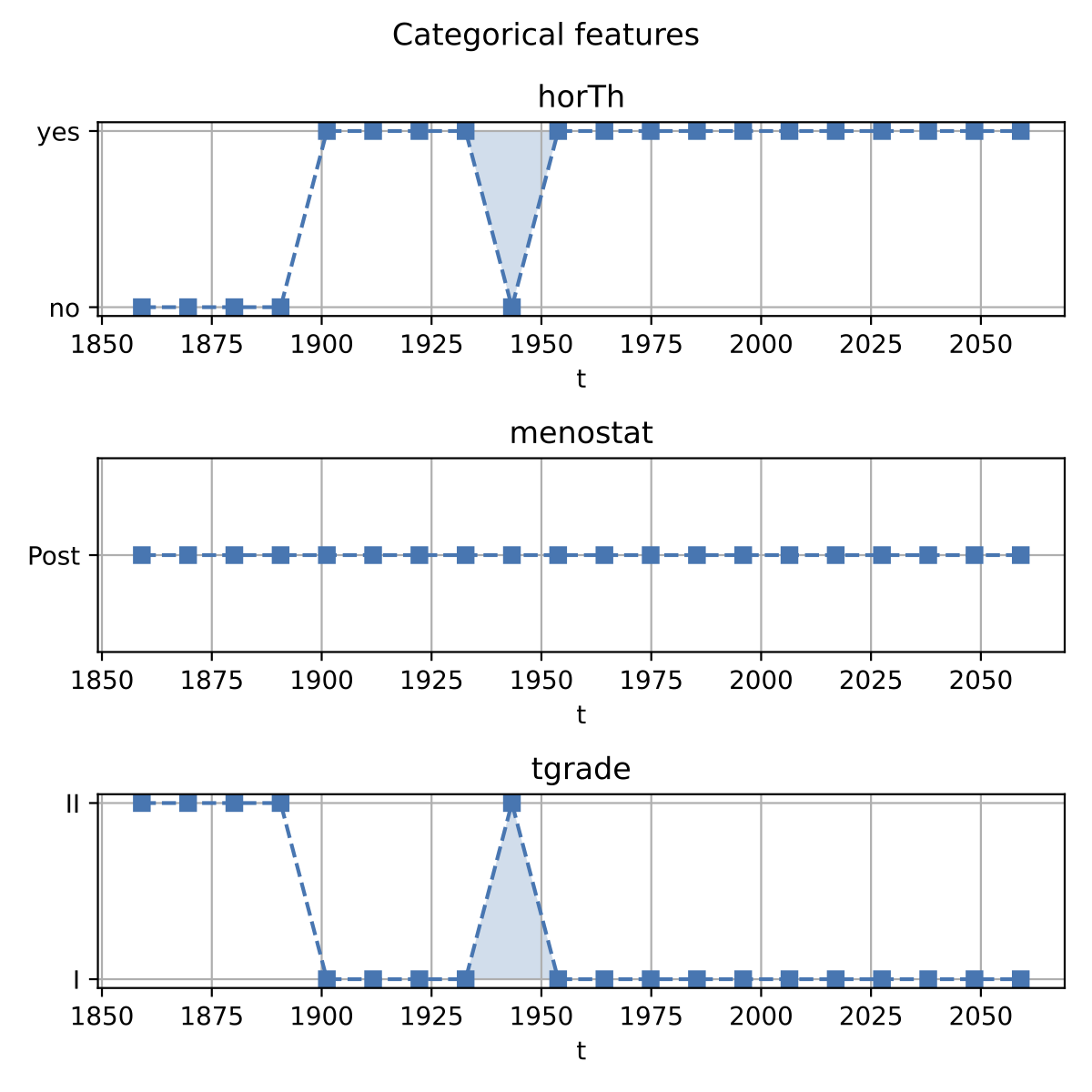}%
\caption{Trajectories of the categorical features generated for the GBSG2
dataset}%
\label{f:gbsg2_categorical_1}%
\end{center}
\end{figure}

The original and generated points are depicted in Fig. \ref{f:gbsg2_rec} by
using the t-SNE method in blue and red colors, respectively. SFs for original
and generated points by using the Kaplan-Meier estimator are shown in Fig.
\ref{f:gbsg2_km}. Trajectories of the continuous features are depicted in Fig.
\ref{f:gbsg2_continuous_1}. In contrast to the previous datasets, where
categorical features do not change in the defined time intervals, the
categorical features of the GBSG2 dataset are changed. This change can be seen
from Fig. \ref{f:gbsg2_categorical_1}. It is interesting to note that all
trajectories have a jump at the same time $1940$. It is likely related to the
unstable behavior of the model with respect to categorical features. Moreover,
it is also interesting to note that changing one feature leads to changes in
all features, indicating strong correlation between features of the considered dataset.%

%

%

%

\subsubsection{Prediction results}

It has been mentioned that the proposed model provides accurate predictions.
In order to compare the model with the Beran estimator \cite{Beran-81}, the
Random Survival Forest \cite{Ishwaran-Kogalur-2007}, and the Cox-Nnet
\cite{Ching-Zhua-Garmire-2018}, we use the C-index. The corresponding results
are shown in Table \ref{t:predict_results}. To evaluate the C-index, we
perform a cross-validation with $100$ repetitions, where in each run, we
randomly select 75\% of data for training and 25\% for testing. Different
values for hyperparameters models have been tested, choosing those leading to
the best results. Hyperparameters of the Random Survival Forest used in
experiments are the following: numbers of trees are $10$, $50$, $100$, $200$;
depths are $3$, $4$, $5$, $6$; the smallest values of instances which fall in
a leaf are one instance, 1\%, 5\%, 10\% of the training instances. Values
$10^{i}$, $i=-3,...,3$, and also values $0.5$, $5$, $50$, $200$, $500$, $700$
of the bandwidth parameter $\tau$ in the Gaussian kernel are selected as
possible values of hyperparameters in the Beran estimator. It can be seen from
Table \ref{t:predict_results} that the proposed model is comparative with the
well-known survival models from the prediction accuracy point of view.%

\begin{table}[tbp] \centering
\caption{Comparison of the proposed model with the Beran estimator, the Random Survival Forest, and the Cox-Nnet for different datasets}%
\begin{tabular}
[c]{ccccc}\hline
Dataset & The proposed model & Beran estimator & Random Survival Forest &
Cox-Nnet\\ \hline
Veteran & $0.711$ & $0.698$ & $0.691$ & $0.707$\\
WHAS500 & $0.758$ & $0.754$ & $0.761$ & $0.763$\\
GBSG2 & $0.679$ & $0.671$ & $0.686$ & $0.672$\\ \hline
\end{tabular}
\label{t:predict_results}%
\end{table}%

\section{Conclusion}

A new generating survival model has been proposed. Its main peculiarity is
that it generates not only additional survival data on the basis of a given
dataset, but also generates the prototype time trajectory characterizing how
features of an object could be changed by different event times of interest.
Let us point out some peculiarities of the proposed model. First of all, the
model extends the class of models which generate survival data, for example,
SurvivalGAN. In contrast to SurvivalGAN \cite{norcliffe2023survivalgan}, the
model is simply trained due to the use of the VAE. The model is flexible. It
can incorporate various survival models for computing the SFs, which are
different from the Beran estimator. The main restriction of the survival
models is their possibility to be incorporated into the end-to-end learning
process. The model generates robust trajectories. The robustness is
implemented by incorporating a specific scheme of weighting the generated
embeddings. The model copes with the complex data structures. It is seen from
numerical experiments where two complex clusters of instances were considered.

In spite of efficiency of the Beran estimator, which takes into account the
feature vector relative location, it requires a specific procedure of
training. Therefore, an idea for further research is to replace the Beran
estimator with a neural network computing the SF in accordance with embeddings
of training data.

We have illustrated the efficiency of the proposed model for tabular data.
However, it can be also adapted to images. In this case, the VAE can be viewed
as the most suitable tool. This adaptation is another direction for research
in future.

\bibliographystyle{unsrt}
\bibliography{Autoencoder,Classif_bib,Explain,MYBIB,Survival_analysis}

\end{document}